\documentclass[journal]{vgtc}                     

\onlineid{0}

\vgtccategory{Research}

\title{InvVis: Large-Scale Data Embedding for Invertible Visualization}

\author{
  Huayuan Ye, Chenhui Li, Yang Li and Changbo Wang
}
\authorfooter{
  \item
  Huayuan Ye, Chenhui Li, Yang Li, and Changbo Wang are with School of Computer Science and Technology, 
  East China Normal University. Chenhui Li is the corresponding author. E-mail: chli@cs.ecnu.edu.cn.
}

\abstract{
  We present InvVis, a new approach for invertible visualization, which is reconstructing or further modifying a visualization from an image. InvVis 
  allows the embedding of a significant amount of data, such as chart data, chart information, source code, etc., into visualization images. 
  The encoded image is perceptually indistinguishable from the original one. We propose a new method to efficiently express chart data in the form of 
  images, enabling large-capacity data embedding. We also outline a model based on the invertible neural network to achieve high-quality data concealing 
  and revealing. We explore and implement a variety of application scenarios of InvVis. Additionally, we conduct a series of evaluation experiments to 
  assess our method from multiple perspectives, including data embedding quality, data restoration accuracy, data encoding capacity, etc. The result of our 
  experiments demonstrates the great potential of InvVis in invertible visualization.
}


\graphicspath{{figs/}{./}} 

\usepackage{tabu}                      
\usepackage{booktabs}                  
\usepackage{lipsum}                    
\usepackage{mwe}                       
\usepackage{amsmath}
\usepackage{algorithm}
\usepackage{algpseudocode}

\usepackage{mathptmx}                  
\usepackage{fancybox}
\usepackage{adjustbox}
\usepackage{marginnote}
\usepackage{fancybox}
\usepackage{multirow}
\usepackage{array}
\usepackage[left=1.2cm,top=1.7cm,bottom=1.6cm,right=1.2cm]{geometry}

\begin{document}
\newcommand{\revision}[1]{{\hypersetup{allcolors=red}\textcolor{red}{#1}}}
\newcommand{\sidecomment}[1]{\marginnote{\adjustbox{minipage=0.43\marginparwidth,fbox}{\textcolor{red}{#1}}}}

\teaser{
  \centering
  \includegraphics[width=\linewidth]{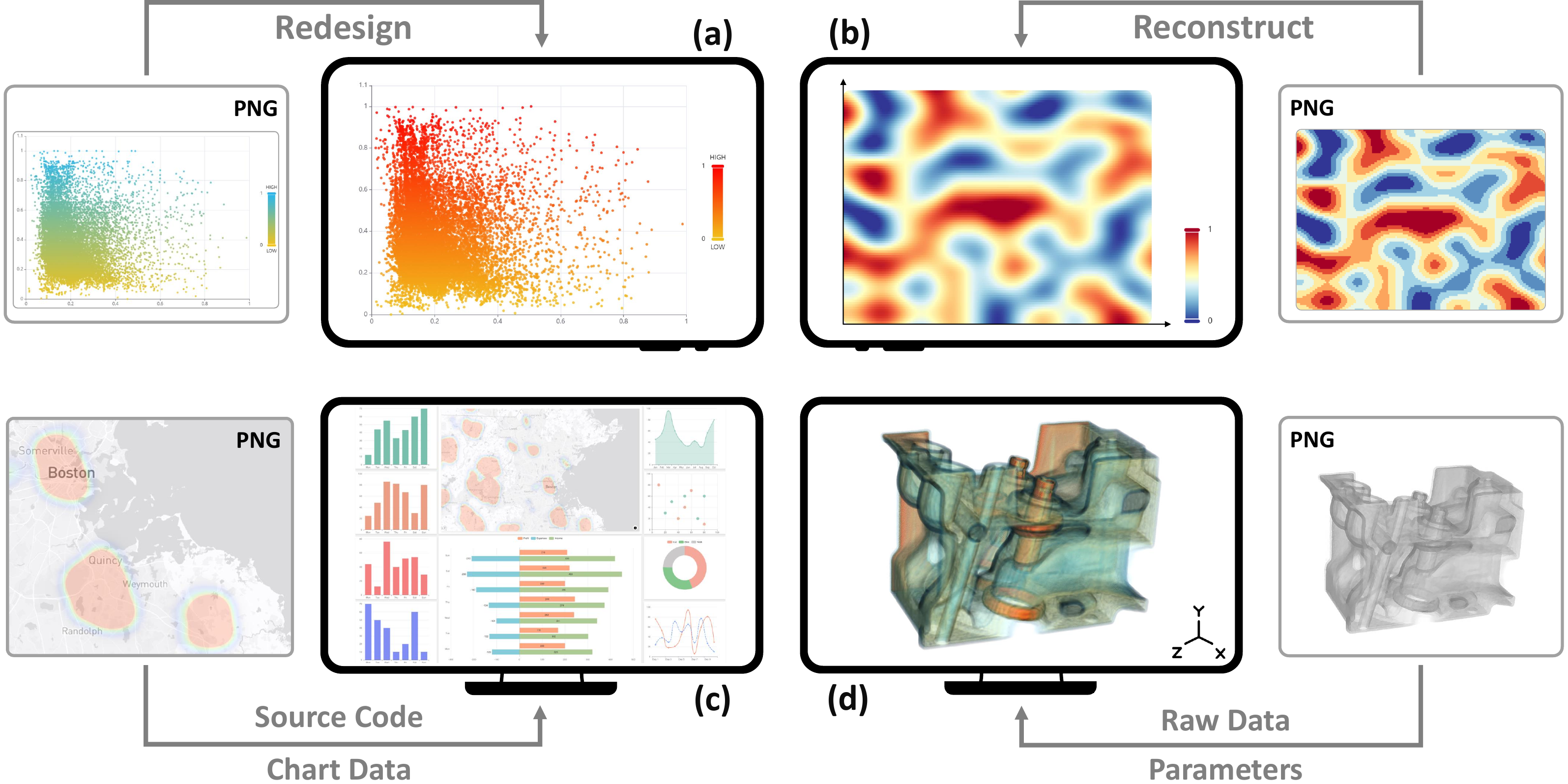}
  \caption{%
    Since InvVis can embed a large amount of data into images, users can decode the embedded data and perform 
    rich exploration, such as redesigning (a) or reconstructing (b) visualizations, rebuilding a visualization dashboard 
    based on the decoded source code and chart data (c), or visualizing volume data based on the decoded raw data 
    and rendering parameters (d).
  }
  \label{fig:teaser}
}


\firstsection{Introduction}
\maketitle

Information visualizations are designed to convey information effectively and efficiently. 
Nowadays, people usually face massive amounts of data. In such cases, the interactivity 
and operability of a well-designed visualization 
allow users to perceive information intuitively and quickly. However, these features are 
often based on certain applications or visualization tools. Typically, the dissemination 
of visualization charts is based on bitmap images. For example, some users may just take a 
screenshot of the visualization chart and share it with other people, this leads to the loss of these 
excellent properties. As a result, it would be meaningful and useful if the initial chart can be recovered
from a chart image. In this paper, we call this procedure as invertible visualization, which is concerned
with reconstructing or further modifying the visualization from an image. 

Some methods have been proposed to address this issue, some of them try to recover the data 
and visual elements using pattern recognition and object detection techniques \cite{poco2017extracting,song2022graphdecoder}. 
However, these methods generally cannot achieve enough recognition accuracy and have limited 
applicability. Some other studies have taken a different approach, which is leveraging 
information steganography, a technique of implicit information embedding. This kind of 
approach \cite{zhang2020viscode, fu2020chartem, fu2022chartstamp} hides
chart information in the image in an inconspicuous way and restores the chart by extracting 
the embedded data. These methods avoid the unstable procedure of identifying chart elements, 
but they cannot achieve a large steganography capacity while ensuring steganography quality, 
making it difficult to deal with charts with large amounts of data. 

Visualizations containing large amounts of data are ubiquitous, such as large-scale scatter 
plots, geographic data visualizations, etc. Actually, it is this kind of visualization charts 
that require more well-designed interactive methods and visual encoding to facilitate users' 
understanding and perception of data. However, previous methods cannot handle the invertible
visualization of this kind of charts due to their limitations mentioned above. Recently, the 
field of information steganography has made significant progress. Many studies have successfully 
hidden arbitrary data or natural images in images without causing significant visual 
differences. Especially, image steganography based on invertible neural network (INN) 
\cite{Dinh2014NICENI, kingma2018glow} has achieved impressive performance. Nevertheless, these methods focus on natural
images instead of chart images and chart data. In this paper, we attempt to address the invertible visualization 
problem of data-intensive charts.

We present a novel method for invertible visualization, called InvVis.
InvVis is an end-to-end pipeline that embeds chart information and chart data into chart images 
in the concealing process and the embedded data can be restored or further reused after the revealing 
process. InvVis can handle the invertible visualization of charts with large amounts of data, which 
is different from previous methods. We propose a data-to-image (DTOI) module which can transfer the
chart data into data images that can be hidden into images in a less perceivable way. We also 
outline a concealing and revealing network based on INN that handles the embedding and restoring of
information. The data images output by the DTOI module, together with the chart 
information that is encoded into QR Code \cite{qrcodeweb} image, are embedded into a chart image through the network. 
And these images can be recovered by the network and transferred back to chart data and chart information. 
The restored data can be used in various scenarios like redesigning the chart, reorganizing chart 
data, etc. 

We conduct a series of experiments and the results show that the encoded chart image generated by 
InvVis have better perceptual similarity compared to other methods, and meanwhile the recovered data 
has less difference. Also, our method can achieve a much larger steganography capacity than previous 
methods. In summary, our InvVis has great potential in addressing the problem of invertible 
visualization, especially in cases involving large amounts of data, where our method can handle the
problem that previous methods cannot. Our contributions include three aspects:

\begin{itemize}
\setlength{\itemsep}{0pt}
\setlength{\parsep}{0pt}
\setlength{\parskip}{0pt}
\item [(1)] {
    We define the problem of invertible visualization that involves large amounts of data. We present 
    various application scenarios of this problem and conduct corresponding explorations and implementations.
}
\item [(2)] {
    We propose a new method of transferring chart data into data images. The data images can effectively represent 
    various kinds of data.
}
\item [(3)] {
    We propose a deep learning-based pipeline to embed information into chart images. Our method can achieve 
    high-quality information concealing and revealing with large embedding capacity.
}
\end{itemize}

\section{Related Work}
\label{sec:rel}

\subsection{Information Steganography}
Information steganography is a technique that conceals information in a carrier with
limited perceptual changes. The carrier can be various formats of data such as image, 
text, audio, video, etc. This technique has considerable applications, e.g., 
digital watermarking, copyright protection and secret communication. An information
steganography method is generally evaluated from three aspects: capacity, security 
and robustness. Capacity means the embedding payload. Security is concerned with the 
undetectability and imperceptibility of embedded data. Robustness refers to the ability 
against distortion. Delina et al. \cite{delina2008information} proposed a text steganography
method to generate steg-text dynamically according to content length and several user-decided 
options. Hota et al. \cite{hota2019embedding} outlined a pipeline to embed 
digital watermarks for scientific visualizations. Yang et al. \cite{yang20163d} embedded
watermarks in a 3D model by modifying its histogram and achieved a fine robustness/distortion 
trade-off. Delforouzi et al. \cite{delforouzi2008adaptive} hid encrypted data into 
the coefficients of audio in the integer wavelet domain while preserving high audio quality.
\cite{swanson1997multiresolution, zhu1999multiresolution} applied information steganography
to carrier videos.

In summary, although previous studies have explored various kinds of carriers, these
methods commonly have limited steganography capacity and are not specially designed for 
information visualization. In our study, a different approach is proposed to address
this problem.

\subsection{Image Steganography}
Image steganography hides data by performing imperceptible changes on a host image.
Traditional methods generally alter the image in spatial domain or transform domain. 
Spatial-domain embedding techniques like Least-Significant Bit (LSB) replacement 
\cite{mielikainen2006lsb} and Bit Plane Complexity Segmentation (BPCS) 
\cite{kawaguchi1999principles} modify pixel values subtly to carry information.
However, these algorithms fail to preserve the statistical properties of images and are
easily detected by steganalysis methods \cite{fridrich2001detecting, yu2004reliable}.
To address this problem, highly undetectable steganography (HUGO) \cite{pevny2010using} 
utilizes high-dimensional image features based on high-order Markov chains to 
improve algorithm security. Transform domain, e.g., discrete cosine transform 
(DCT) domain \cite{almohammad2008high} and discrete wavelet-transform (DWT) domain 
\cite{swanson1997multiresolution, zhu1999multiresolution}, can also be modfied to 
embed information. Although transform-domain embedding is generally more robust
than spatial-domain embedding, both of them can only hide limited bit-level information.

Recently, a variety of deep learning-based image steganography schemes have been 
proposed and have achieved impressive performance. 
\cite{zhu2018hidden,tancik2020stegastamp} adopted the autoencoder to embed 
binary messages in a cover image. \cite{zhang2019steganogan, qin2020coverless} 
used generative adversarial network (GAN) to optimize the distortion of encoded 
images. There were also various studies that attempted to hide one or more secret 
images into a carrier image. Baluja et al. \cite{baluja2017hiding} first utilized 
an end-to-end network to learn feature representations of color images and 
embed them into cover images. Wengrowski et al. \cite{wengrowski2019light} introduced
a photographic steganography algorithm based on the encoder-decoder network for light 
field messaging (LFM). Wu et al. \cite{wu2021embedding} proposed to embed multiplane 
images (MPIs) in a JPEG image and synthesize novel views by decoding it.
More recently, the invertible neural network (INN) \cite{kingma2018glow, Dinh2014NICENI} has been 
utilized to hide single \cite{jing2021hinet, lu2021large} or multiple 
\cite{cheng2021iicnet, guan2022deepmih} images into carrier images. 

Most of the aforementioned image steganography methods are designed to hide information 
in natural images. However, we focus on embedding data into chart images. Chart images 
typically comprise many homogeneous regions and have different features from natural images.
As a result, a different method is required to handle this problem.

\begin{figure*}[htb]
\begin{minipage}{\dimexpr\linewidth-4\fboxsep-2\fboxrule\relax}
    \centering
    \includegraphics[width=1.0\linewidth]{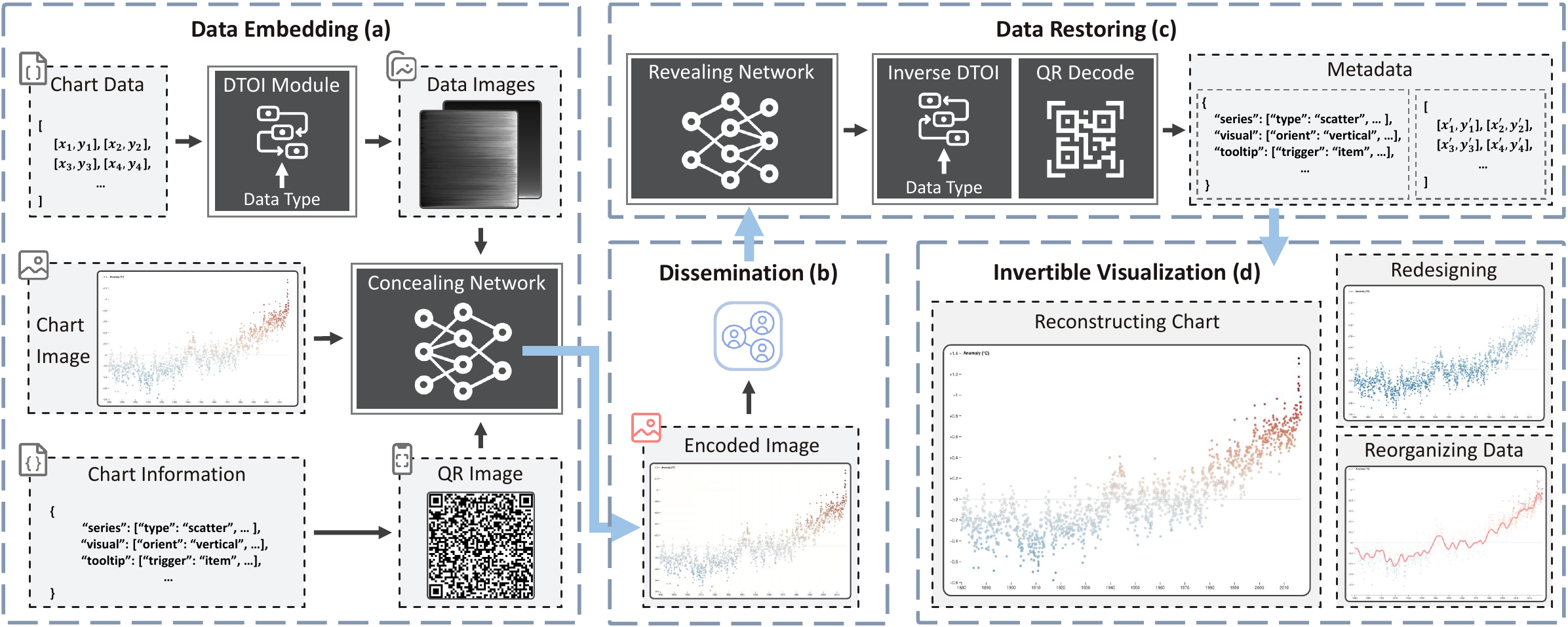}
    \caption{
        An overview of our InvVis pipeline. In the data embedding process (a), the user first uploads a 
        chart image as the carrier to embed data. Then, the metadata of the chart, including
        the chart data and chart information required for invertible visualization, will be uploaded 
        and processed into data images and QR Code image separately. Next, the concealing network 
        takes the abovementioned images as input and generates the encoded image. The encoded image can 
        be disseminated (b). The receiver may view it as a normal image or decode it with the revealing 
        network and restore the metadata (c). The recovered data can be used for invertible visualization (d), 
        which facilitates various scenarios like reconstructing or redesigning the chart, reorganizing chart data, etc.
    }
    \label{fig:pipline}
\end{minipage}
\vspace{-10pt}
\end{figure*}

\subsection{Invertible Visualization}
Invertible visualization allows users to regain or further modify the visualization chart
from a raster image. The reconstructed chart preserves its initial information 
including chart type, chart data, interactive methods, visual mapping, etc.
This facilitates scenarios such as redesigning the chart, reorganizing 
chart data, helping visually impaired users to understand chart
\cite{choi2019visualizing, singh2021chartsight, poco2017reverse}, 
and enabling reproducibility
for scientific visualizations \cite{10.1145/1142473.1142574}.

Previous studies on invertible visualization mainly took two kinds of approaches.
First kind of methods aim to extract the information directly from the chart image.
Savva et al. \cite{savva2011revision} used machine learning techniques to 
infer chart type and underlying data. Poco et al. \cite{poco2017reverse} proposed a 
pipeline to automatically recover the visual encoding specification of a chart from 
a raster image. \cite{poco2017extracting} utilized a convolutional neural network (CNN) to
extract color mappings from visualization images. Later, some studies tried to
improve the extraction performance by introducing human interactions
\cite{mendez2016ivolver, flower2016validity} or focusing on certain types of charts 
\cite{al2017machine, Liu2019DataEF, song2022graphdecoder, chen2019towards}.
However, this kind of methods are still likely to suffer from insufficient 
accuracy or limited application scenarios due to the diversity and structural 
complexity of visualization charts. Another kind of methods embed the metadata of
charts in the generated visualization images and reconstruct the chart by
decoding the image. VisCode \cite{zhang2020viscode} trains an encoder-decoder 
network to embed and recover QR Codes which contain chart information. To optimize 
the encoding quality, VisCode leverages a visual importance network to guide the 
embedding process. Chartem \cite{fu2020chartem} detects the background of chart images and 
hides metadata by slightly modifying background pixel values. ChartStamp 
\cite{fu2022chartstamp} utilizes a distortion layer that simulates real-world image 
manipulations to improve the robustness. This kind of methods have addressed the 
abovementioned problem to some extent at a cost of the perceptual quality of chart images. 
However, to our best knowledge, few previous study has addressed the issue
of invertible visualization when the data volume is large. In this paper,
we mainly focus on embedding and restoring large amounts of data in the 
context of information visualization, and we present a new method to implement 
invertible visualization.

\section{Overview}
\label{sec:ove}

In the field of information visualization, chart plays a critical role in describing
and presenting data. Ideally, a well-designed chart should possess 
properties like user-friendly interactive methods and reasonable demonstration
of chart data. These features can improve the efficiency and accuracy for users to 
cognize and comprehend data, even when the data volume is large at times. Unfortunately,
in most cases, visualization charts are disseminated in the form of raster images,
forfeiting the aforementioned features. This greatly limits users' ability to 
explore and interact with the chart. To solve this problem, we aim to implement
invertible visualization by encoding the
metadata of the chart into the generated chart image using image steganography
techniques. 

In this paper, we propose InvVis, a new approach for invertible visualization.
Different from previous studies \cite{zhang2020viscode, fu2020chartem, fu2022chartstamp}, 
we focus not just on visualization charts with relatively limited amounts of metadata, 
but also on those with large data volume. \autoref{fig:pipline} shows the pipline
of our InvVis, which contains three main components:

\begin{itemize}

\item \emph{Data-to-Image Module}
To ensure that the chart data enters the concealing network in the form of images, we propose
a data-to-image (DTOI) module to implement the data-to-image transfer.
This module takes chart data as input and generates several data images representing the data
based on the data type. In addition, this module is invertible, which means the original data can be restored by performing the 
inverse operation of the DTOI module on these images. 

\item \emph{Concealing Network}
The concealing network embeds the metadata of the chart in the carrier image. At first,
data images and QR Code image are padded to the same size as the carrier image.
Then, these images, including the carrier image, are concatenated and fed into
the concealing network. The network produces an encoded image with the same 
resolution as the carrier image while introducing subtle perturbations.

\item \emph{Revealing Network}
The revealing network retrieves the metadata from an encoded image. The network
first decodes an encoded image to obtain the restored data images and QR Code image.
After performing corresponding operations on these images, the restored chart data 
and chart information are available to users. Users may use the restored data for 
various applications (e.g., redesigning the chart). 
    
\end{itemize}

\begin{figure*}[htb]
    \centering
    \includegraphics[width=0.9\linewidth]{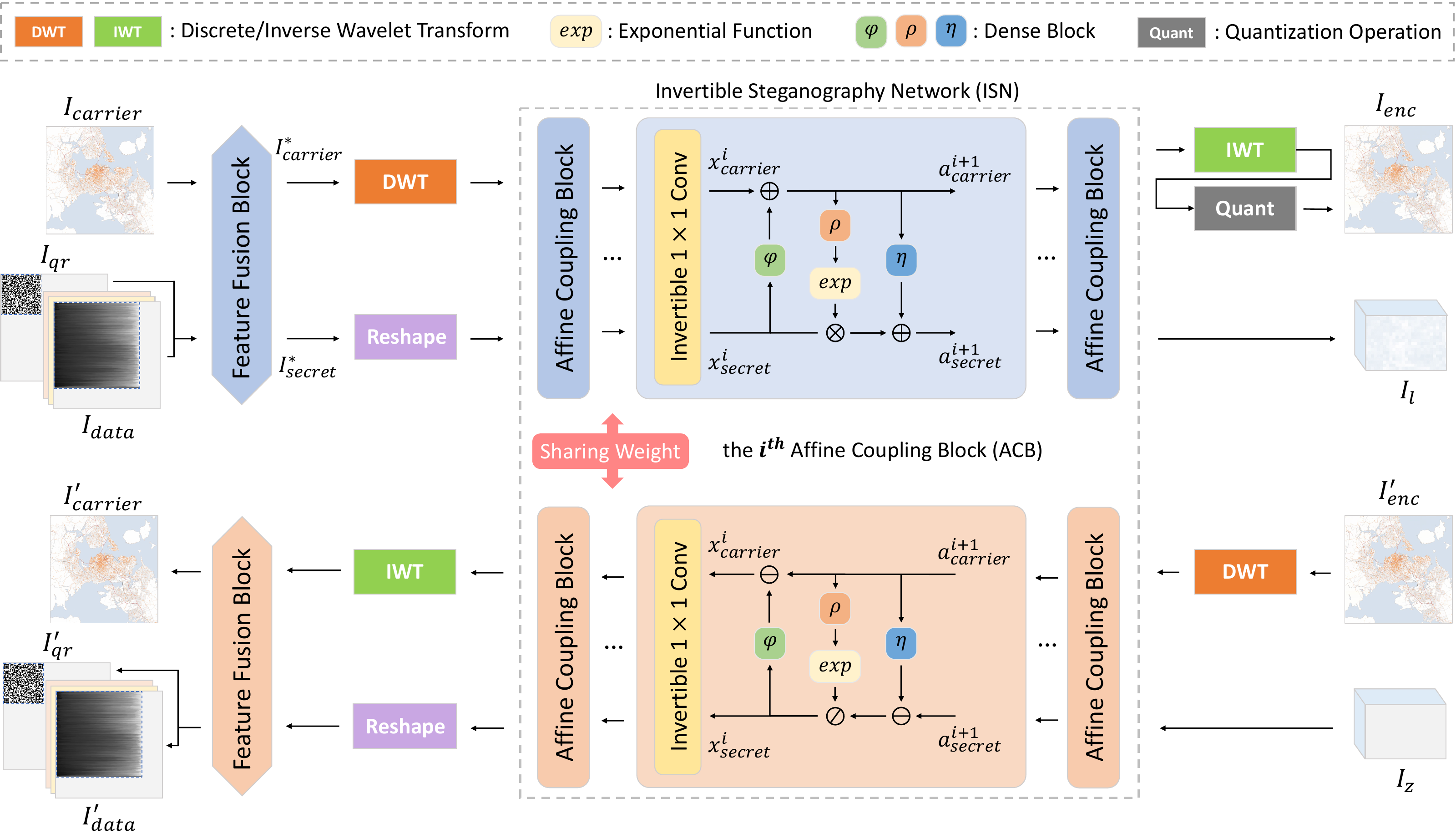}
    \caption{
        The architecture of our network. The upper part is the concealing network, and the lower part is the revealing network. The concealing
        network takes a carrier image and several secret images as input and outputs an encoded image, while the revealing network performs
        the opposite operation.
    }\vspace{-15pt}
    \label{fig:flowmodel}
\end{figure*}

\subsection{Definition}
In our InvVis, the input comprises three parts, a carrier image $I_{c}$, 
a sequence of chart data $S$, and a plain text $T$ that represents the 
chart information. The DTOI module $Dtoi(\cdot)$ transfers the chart 
data to $K$ data images $I_{d}^{1:K}$. The chart information is encoded into
several QR Codes and stitched together into a QR Code image $I_{qr}$. These images
are fed into the concealing network $E(\cdot)$ and result in an encoded image $I_{e}$. 
In the revealing process, the revealing network $D(\cdot)$ takes an encoded image 
$I^{'}_{e}$ as input and outputs the restored data images $I_{d}^{'1:K}$ and QR Code 
image $I^{'}_{qr}$. After that, the restored chart data $S^{'}_{d}$ and chart information 
$T^{'}$ can be obtained by performing the reverse DTOI operation $Dtoi^{-1}(\cdot)$ and QR Code 
decoding operation separately. For our InvVis, we aim to make the encoded image perceptually 
similar to the original chart image, while ensuring that the decoded data is as accurate as 
possible. Formally, we aim to optimize the two networks $E(\cdot)$ and $D(\cdot)$ by minimizing:\vspace{-5pt}
\begin{equation}
    \left\| I_{c} - I_{e} \right\| + 
    \lambda _1 \sum_{k=1}^{K} \left\| I_{d}^{k} - I_{d}^{'k} \right\| +
    \lambda _2 \left\| I_{qr} - I_{qr}^{'} \right\|
    ,\vspace{-7pt}
\end{equation}
where $\lambda _1$ and $\lambda _2$ are weight coefficients.

\section{Methods}

\subsection{Dataset}
\label{sec:dataset}
Our dataset can be divided into three parts, which are visualization images,
data images and QR Code images.

\noindent\textbf{Visualization Images} Visualization images are commonly composed of 
homogeneous regions, which results in a significant difference from natural 
images and therefore leads to different image features. Hence, it is not suitable to 
use a natural image dataset to train our model. Instead, we used a subset of images
from VIS30K \cite{Chen:2021:VCF}, which is a dataset containing visualizations 
from various sources, as our visualization image dataset. Specifically, we selected 
a total of 1500 images and split them into 1200 images as training set and the rest 
300 images as testing set.

\noindent\textbf{Data Images} We trained our model using data images as input instead 
of raw chart data. We built our data image dataset with two kinds of data, discrete data 
and continuous data. For discrete data, we randomly constructed 300 sets of 2-dimensional 
scatter data. And for continuous data, we first constructed 300 Perlin noise 
images \cite{perlin1985image}, which are spatial-continuous. To make our dataset more 
comprehensive, we also selected 380 sets of data from the Earthdata website 
\cite{EarthScienceDataSystems}, which include wind field data, ocean current data and 
other kinds of data with continuity. All types of data mentioned above were transferred 
into data images using the DTOI module to be discussed in \ref{sec: dtoi}. We
divided these data images into training and testing sets in a 4:1 ratio.

\noindent\textbf{QR Code Images} For the purpose of invertible visualization,
the chart information is required to be restored from the encoded image with no error.
Thus, it is necessary to use a reliable coding scheme to encode this kind of information.
There are many coding schemes with error correction (ECC) ability, like Polynomial Code
\cite{Moore2017PolynomialCO}, BCH Code \cite{Bose1960OnAC}, etc. However, embedding such kind 
of binary data can cause perceptible artifacts in the encoded image and also lead to insufficient 
data recovery accuracy, which is not suitable for the invertible visualization problem.
As a result, we leverage QR Code \cite{qrcodeweb}, which is a coding scheme with 
error correction. Although the encoding capacity of QR Code is 
less than the abovementioned coding schemes,
it is easier to embed and can achieve higer data recovery accuracy.

QR Code has different symbol versions (Version 1 to Version 40), and 
it provides four ECC levels, higher ECC level means higher error correction rate.
For more details, we suggest that the reader refer to the specification of 
ISO/IEC 18004 \cite{qrspec}.
In our implementation, we utilize QR Code Version 40 to encode chart information, 
whose information capacity and error correction capability for different ECC levels is shown in 
\autoref{tab:qrecc}. In addition, we choose the `H' ECC level, which means about 
30\% of the error can be fixed. 
For the training data, we constructed 500 random strings whose length 
range from 1 to 1273 and converted them into QR Code images.\vspace{-5pt}

    
    

\begin{table}[htb]
    \caption{Properties of QR Code Version 40}
    \vspace{-7pt}
    \newcolumntype{M}[1]{>{\centering\arraybackslash}m{#1}}
    \renewcommand\arraystretch{1.2}
    \centering
    \small
    \begin{tabular}{@{}M{2.6cm}M|M{2.6cm}M{2.6cm}|@{}}
    \bottomrule
    \multicolumn{2}{|c|}{ECC Level} & {Information Capacity} & {ECC Capability} \\ [0.5pt]
    \hline
    \multicolumn{2}{|c|}{`L' (`Low')} & {2953 characters} & {About 7\%}  \\
    \multicolumn{2}{|c|}{`M' (`Medium')} & {2331 characters} & {About 15\%}  \\
    \multicolumn{2}{|c|}{`Q' (`Quality')} & {1663 characters} & {About 25\%}\\
    \multicolumn{2}{|c|}{`H' (`High')} & {1273 characters} & {About 30\%}\\ [0.5pt]
    \toprule
    \end{tabular}\vspace{-10pt}
    \label{tab:qrecc}
\end{table}

\subsection{Data-to-Image Module}
\label{sec: dtoi}
Previous research has proposed various methods to hide data in images. 
\cite{zhang2020viscode, wengrowski2019light} hid QR Codes into images, while 
\cite{tancik2020stegastamp, fu2020chartem, fu2022chartstamp,zhu2018hidden}
directly embedded binary data into images.
However, these methods can only achieve limited data embedding capacity. 
To solve this problem, we propose a data-to-image (DTOI) module, which is a new data coding 
scheme. The data images output by the DTOI module can be embedded into the carrier 
image with a larger capacity while bringing less perceptual distortion than 
previous methods. 

In terms of data visualization, most data can be classified into two types: 
continuous data and discrete data. Our DTOI module performs different operations on
these two kinds of data, this process will be described in detail in the following subsections.

\begin{figure}[htb]
    \centering
    \subfloat[Data image of a heat map]
    {\includegraphics[width=1.0\linewidth]{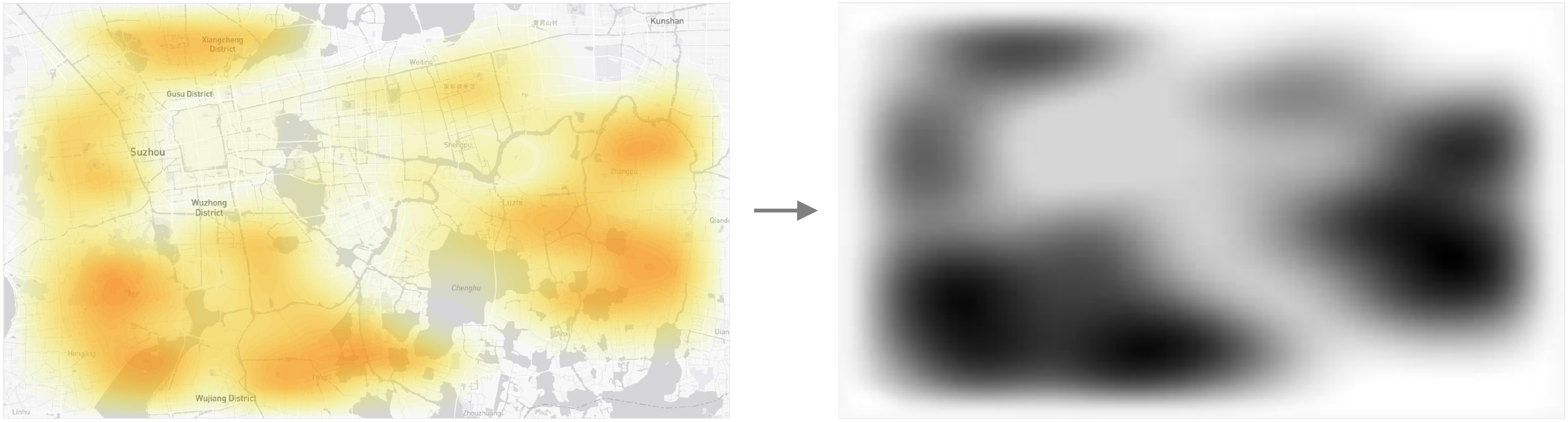}}
    \hfil
    \subfloat[Data images of a volumetric data visualization]
    {\includegraphics[width=1.0\linewidth]{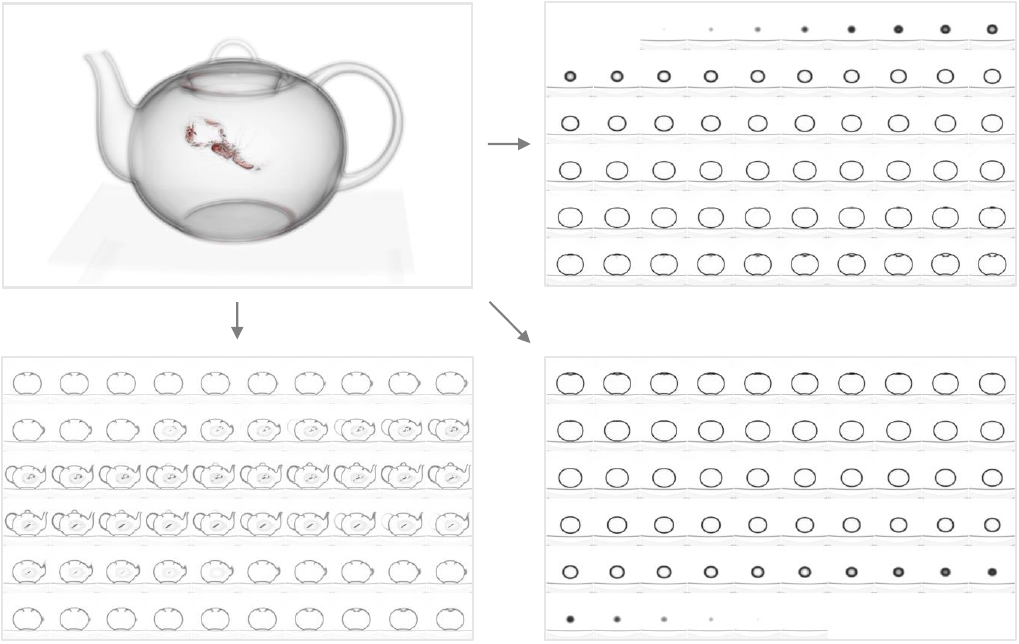}}
    \hfil\vspace{-7pt}
    \caption{\label{fig:di_c}Continous data images output by the DTOI module.}\vspace{-15pt}
\end{figure}

\subsubsection{DTOI for Continuous Data}
Within the realm of data visualization, there are many kinds of continuous data that 
exhibit spatial continuity in some dimensions, for example, density maps. Although some 
methods obtain density maps by performing kernel density estimation (KDE) on scatter 
data, for some data that contains large amounts of scatter points, this is quite 
time-consuming. In such cases, directly storing the entire density map as a grayscale
image can also be a choice.
There are many other types of continuous data, vector field data such as wind field and 
flow field data have spatial continuity in the components of the vectors in each dimension.
Volumetric data is another example, it has spatial continuity on every cross-section.

Formally, a sequence of continuous data can be represented as $P=\{ p_i \}^{N}_{i=1}$,
where $p_i$ is a 2-dimensional plane of spatial-continuous data and $N$ is the number 
of such planes. For instance, $N$ is 1 for a density map and 2 for 2-dimensional wind 
field data. Due to the varying distribution and range of 
different chart data, we normalize $P$ as follows:

\vspace{-5pt}
\begin{equation}\label{eq: norm}
    \hat{P} = \{ \frac{p_i - m_i}{M_i - m_i} \}^{N}_{i=1}
    ,\vspace{-5pt}
\end{equation}
where $\hat{P}$ is the normalized $P$ while $m_i$ and $M_i$ are the minimum and maximum 
values in $p_i$, respectively. After that, we convert each element in $\hat{P}$ to a 
single-channel grayscale image and obtain $N$ data images. In the case when $N$ is large 
(e.g., volumetric data), these data images will be stitched horizontally 
and vertically to minimize the number of data images, while ensuring that the size of 
the stitched image does not exceed that of the carrier image.
Note that all $m_i$ and $M_i$, together with the data image stitching information are encoded 
into QR codes, just like chart information, since the normalized data needs to be 
de-normalized during the data restoration process.

\autoref{fig:di_c} shows some continuous data images output by the DTOI module.
\autoref{fig:di_c}(a) is a heatmap visualization and its corresponding data image, 
which is a density map. \autoref{fig:di_c}(b) shows a volumetric data visualization,
whose data images are stitched together to reduce their number.

\subsubsection{DTOI for Discrete Data}\label{sec:dtoi_discrete}
Discrete data is also very common in the field of information visualization, e.g., 2-dimensional scatter data. 
For some other kinds of data with higher dimensions, methods such as t-SNE \cite{van2008visualizing} 
or UMAP \cite{2018arXivUMAP} are often used to reduce their dimensionality for easier visualization. 
Most of this kind of data becomes 2-dimensional after dimensionality reduction, thus, 
the visualization of 2-dimensional discrete data is very common and important. In this 
paper, we focus on 2-dimensional discrete data.
This kind of data generally lacks continuity, either 
within or between dimensions. If we treat it the same way as we treat continuous data, 
the result data images will also lack continuity and features, making it difficult for the network to learn 
feature representations, which in turn leads to unsatisfactory results in data concealing 
and revealing. As a result, we utilize a novel method to transform discrete data into relatively 
continuous and smooth data images to solve this problem.

\begin{figure}[htb]
    \centering
    {\includegraphics[width=1.0\linewidth]{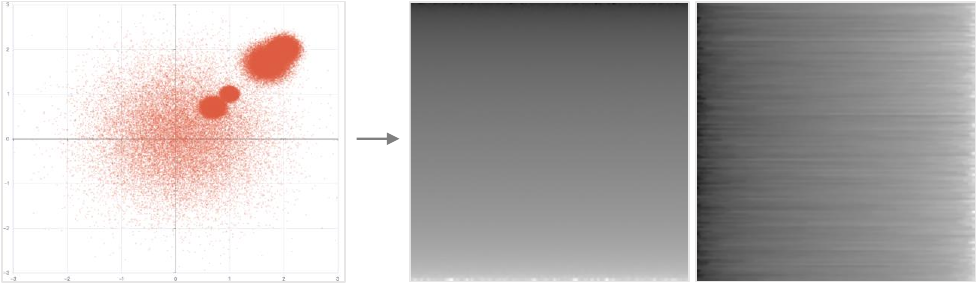}}
    \caption{\label{fig:dtoi_d_scatter}
        The data images correspond to the scatter plot.
    }\vspace{-15pt}
\end{figure}

Given a set of 2-dimensional discrete data $S = \{(x_i, y_i)\}^N_{i = 1}$ where 
$N$ is the number of tuples in $S$, we first sort it in ascending order with the first 
dimension as key, which is, for $(x_i, y_i)$ and $(x_j, y_j)$ in $S$, the former is considered 
to be larger then the latter if $x_i > x_j$. Then, we divide $S$ into groups and sort each group 
locally with the second dimension as key. This promises that the first dimension data of $S$ are 
close to each other and the second dimension data are in ascending order in each group. This endows 
the originally discrete and irregular data with a certain continuity.
$S$ is then normalized and reshaped to obtain 2 grayscale images.
To further enhance the continuity of the data images, we insert $K$ additional pixels 
between every two adjacent pixels and use linear interpolation to assign values to them.
Specifically, for every pair of adjacent pixels $p_a$ and $p_b$, the inserted pixel values are:

\vspace{-5pt}
\begin{equation}\label{eq:interp}
    p_i = \frac{K - i + 1}{K + 1}p_a + \frac{i}{K + 1}p_b, \ i = 1, 2 ... K
    ,\vspace{-5pt}
\end{equation}
where $p_i$ is the value of the $i$\textsuperscript{th} inserted pixel and $K$ 
is set to 3 in this paper.
Larger $K$ can enhance the continuity of data images but decrease the steganography capacity.
We also conduct an experiment on how the value of $K$ affects the 
data embedding quality, this will be discussed in \ref{sec:steg_q}.
This algorithm is explained in detail in Algorithm \ref{alg:dtoi_d}. In addition, if the
data volume is quite large, the data can be divided into parts and apply this algorithm 
respectively to generate several data images. These images can then be stitched
together to reduce the total channel number of the final data images. \autoref{fig:dtoi_d_scatter} shows
the 2 data images generated with our algorithm, as we can see, the data images are smooth and continuous.
In the data restoring procedure, with the revealed data images output by the revealing 
network, the restored data can be obtained by selecting and denormalizing those non-inserted pixel values. 

\begin{figure}[htb]
    \centering
    \begin{minipage}[t]{0.48\linewidth}
    \centering
    \includegraphics[width=\linewidth]{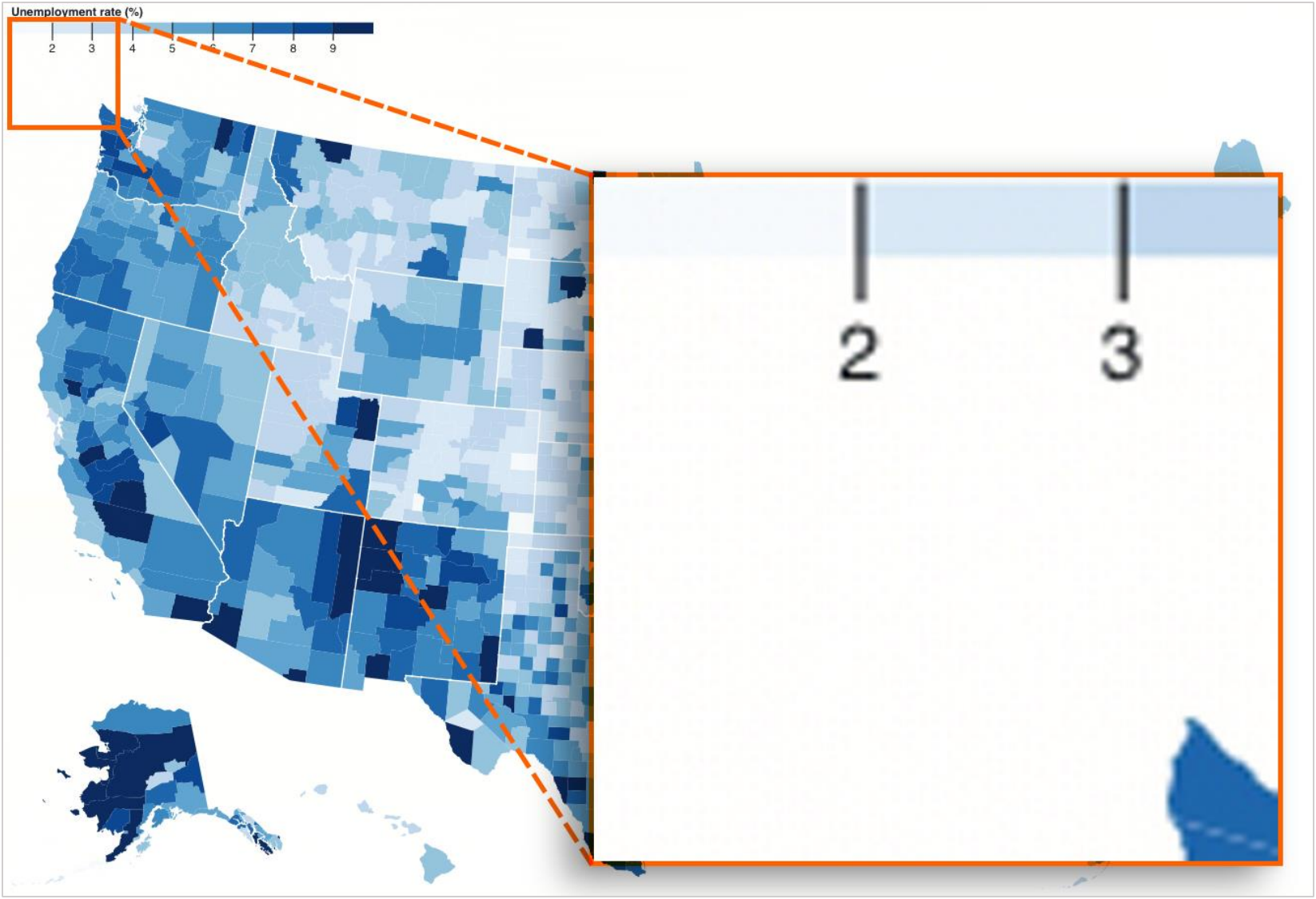}
    \subcaption{Using our algorithm}
    \end{minipage}\hfill
    \begin{minipage}[t]{0.48\linewidth}
    \centering
    \includegraphics[width=\linewidth]{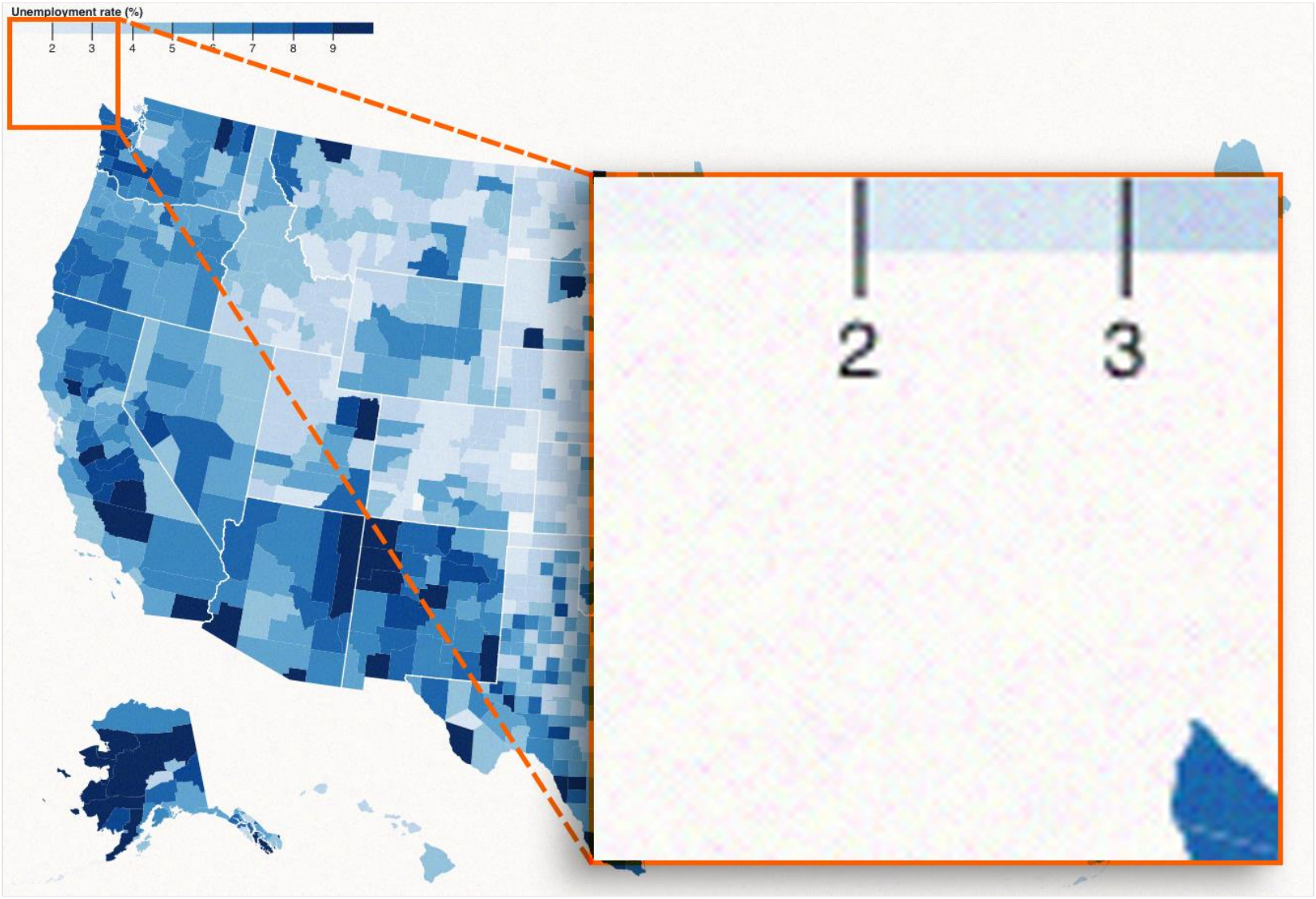}
    \subcaption{Not using our algorithm}
    \end{minipage}\vspace{-8pt}
    \caption{\label{fig:dtoi_d_compare}Embedding result of using and not using our algorithm.}\vspace{-15pt}
\end{figure}

As shown in \autoref{fig:dtoi_d_compare}, we first embed data images obtained using the above algorithm 
into a chart image and result in \autoref{fig:dtoi_d_compare}(a). For comparison, we also embed data 
images obtained without sorting and pixel insertion operation into the same chart image and result in 
\autoref{fig:dtoi_d_compare}(b). As we can see, the data images generated with our algorithm can be embedded
into images with less distortion.

\begin{algorithm}[htb]
    \caption{DTOI Algorithm for Discrete Data}
    \label{alg:dtoi_d}
    \begin{algorithmic}[1]
        \Require
            $S = \{(x_i, y_i)\}^N_{i = 1}$: a set of 2-dimensional discrete data containing $N$ elements; 
            $(H_s, W_s)$: two integers close to each other and $H_s \times W_s \ge N$;
            $sort(s, i)$: a function that sort the array $s$ in ascending order with its 
            $i$\textsuperscript{th} dimension as key; $K$: the number of additional pixels inserted between two adjacent pixels;
        \Ensure
            $I_{data}$: a set of result data images, where
            the size of each image is $((K + 1) \times H_s, (K + 1) \times W_s)$;

        \State sort($S$, 1)
        \State $s_p \leftarrow the \ last \ element \ of \ S$
        \While {$S$.size < $H_s \times W_s$}
        \State $S$.append($s_p$)
        \EndWhile
        \State $S \leftarrow divide \ into \ H_s \ groups \ equally$
        \For {$s_{group}$ \textbf{in} $S$}
        \State sort($s_{gruop}$, 2)
        \EndFor
        \State $\hat{S} \leftarrow normalize \ S \ with \ \autoref{eq: norm}$
        \State $I_{data} \leftarrow reshape \ \hat{S} \ to \ (2, H_s, W_s) \ and \ convert \ 
        to \ 2 \ grayscale \ images$
        \For {$I_d$ \textbf{in} $I_{data}$}
        \For {\textbf{each} $row$ \textbf{in} $I_d$}
        \State $row \leftarrow insert \ K \ pixels \ with \ \autoref{eq:interp}$
        \EndFor
        \For {\textbf{each} $column$ \textbf{in} $I_d$}
        \State $column \leftarrow insert \ K \ pixels \ with \ \autoref{eq:interp}$
        \EndFor
        \EndFor
        \State\Return $I_{data}$
    \end{algorithmic}
\end{algorithm}

\vspace{-2pt}

\subsection{Concealing and Revealing Network}
\autoref{fig:flowmodel} shows our concealing and revealing network, which is a 
symmetric architecture. Both of these two networks are composed of a feature 
fusion block (FFB) and an invertible steganography network (ISN). In the concealing 
process, the input carrier image $I_{carrier}$ and secret images $I_{secret}$ which 
consist of QR Code image $I_{qr}$ and data images $I_{data}$, are fed into the FFB.
To minimize the impact of QR Code image on the distortion of encoded images, we multiply the QR
code image by a coefficient $m_{qr}=0.15$ before feeding it into the network, and the 
recovered QR code image is divided by $m_{qr}$ during the decoding procedure.
The output of FFB, i.e. $I^{*}_{carrier}$ and $I^{*}_{secret}$, undergo discrete wavelet
transform (DWT) and reshaping separately, they are then concatenated as the input of ISN.
The ISN processes its input and results in an encoded image $I_{enc}$ after performing
the inverse wavelet transform (IWT) and quantization operation. Since the input and output 
sizes of ISN are the same, it also outputs a matrix $I_l$. 
In the decoding process, an encoded image $I^{'}_{enc}$, together 
with a constant matrix $I_z$ whose size is the same as that of $I_l$, goes through the ISN
and FFB in sequence to produce the restored data images $I^{'}_{data}$ and QR Code image 
$I^{'}_{qr}$.  Note that in our implementation, the maximum number of the channel that hides
information is 4 (3 channels for data images and 1 channel for QR Code image), but it is 
quite easy to change by slightly modifying the network.

\subsubsection{Feature Fusion Block}
Previous studies \cite{guan2022deepmih,jing2021hinet,cheng2021iicnet,lu2021large} have 
proposed some methods to hide natural images in each other. The secret images and
carrier images in these methods generally have similar natural image features. However, 
in our InvVis, the secret images to be embedded are data images and QR Code image, they 
have different data distributions and features from the carrier image. Hence, a different
method is needed to reduce the perceptual deviation of the encoded image. To address 
this issue, we propose the feature fusion block (FFB), attempting to optimize the 
perceptual quality by blending the features of the carrier image and secret images before
the steganography process.

\begin{figure}[htb]
    \centering
    {\includegraphics[width=0.9\linewidth]{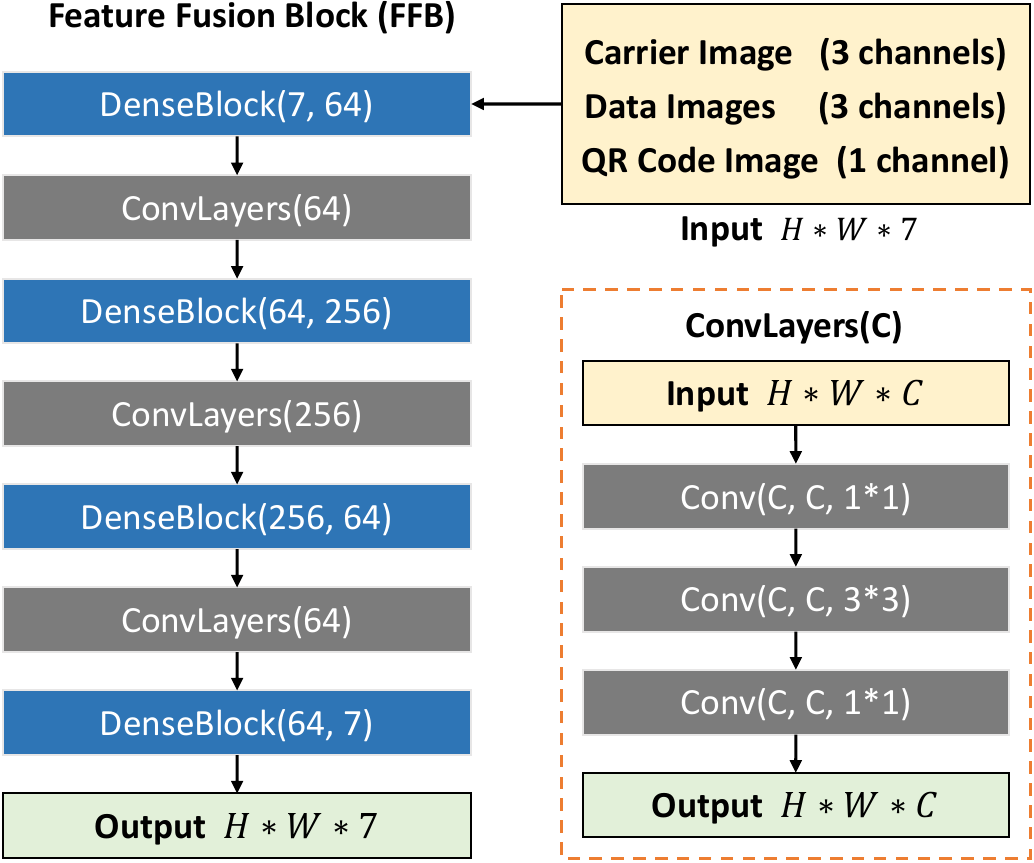}}
    \caption{\label{fig:ffb}
        The architecture of our feature fusion block (FFB).
    }\vspace{-17pt}
\end{figure}

The architecture of our FFB is shown in \autoref{fig:ffb}, which is composed of several
dense blocks \cite{wang2018esrgan} and common convolutional layers. The input of FFB
comprises a total of 7 channels, and its output has the same size as the input. In the
concealing process, the FFB projects the input into a high-dimensional space and 
then maps it back to a low-dimensional space, achieving the effect of feature blending.
The output of FFB is easier to result in an encoded image with less visual distortion in 
the following steganography network. The ablation experiment on FFB will be discussed 
in \ref{sec:steg_q}. In the decoding process, we also use an FFB to process the output of 
ISN and obtain the final restored images.

\subsubsection{Wavelet Domain Steganography}
Spacial-domain steganography can lead to texture-copying artifacts 
\cite{weng2019high}, which affects the perceptual quality and is easy to be detected.
To improve the concealing performance, we adopt an approach similar to HiNet
\cite{jing2021hinet} and DeepMIH \cite{guan2022deepmih}, which is introducing discrete 
wavelet transform (DWT). Transferring the image to the wavelet domain before the ISN can 
guide the network to learn a better embedding strategy that hides more secret information 
in the high-frequency part of the image rather than the low-frequency part.

Specifically, we perform DWT on $I^*_{carrier}$, which is part of the output of FFB 
corresponding to the initial carrier image $I_{carrier}$. Assume that the size of 
$I^*_{carrier}$ is $(C_c, H, W)$ where $C_c$, $H$, $W$ indicate the channel number, 
height and width of the image, DWT transfers its size to $(4C_c, H / 2, W / 2)$. 
In contrast, we do not perform this operation on $I^*_{secret}$ whose size is 
$(C_s, H, W)$, instead, we simply reshape it to $(4C_s, H / 2, W / 2)$ for the purpose 
of size matching. After the hiding process performed by ISN, we use inverse wavelet
transform (IWT) to transform the image back to spatial domain and obtain the encoded image.
Similarly, In the revealing procedure, a series of symmetric operations are performed 
to restore the secret images. We conduct the ablation experiment and the result shows that this
method can lead to better steganography quality, this will be discussed in \ref{sec:steg_q}.

\subsubsection{Invertible Steganography Network}
Our invertible steganography network (ISN) is based on the invertible neural network
(INN), which was first proposed by Dinh et al. \cite{Dinh2014NICENI}, it allows the
secret information to be hidden and revealed in the same network. To enhance the 
representation ability of the network, we also leverage the invertible 1 $\times$ 1 
convolution block proposed by Kingma et al. \cite{kingma2018glow}. Formally, given $c$ 
and $s$, which are the input of ISN corresponding to the carrier image and secret images, 
the encoded image $e$ can be obtained with $e = f_{\theta}(c, s)$ where $f(\cdot)$ is the 
hiding process and $\theta$ indicates the network parameters. For the revealing 
procedure, an inverse process $f^{-1}_{\theta}(\cdot)$ which shares the same parameters 
$\theta$ as $f_{\theta}(\cdot)$ can output the restored images $c'$ and $s'$ with 
$(c', s') = f^{-1}_{\theta}(e)$.

As shown in \autoref{fig:flowmodel}, our ISN is composed of several affine coupling blocks
(ACBs). Each ACB shares the same parameters in the hiding and revealing process but has
an opposite data flow direction. 
Assume that the input of the $i$\textsuperscript{th} ACB is $a^i_{carrier}$ and
$a^i_{secret}$ indicating the carrier image and secret image channels, the invertible 
convolution block outputs $x^i_{carrier}$ and $x^i_{secret}$ with their size not changed:
\begin{equation}\label{eq:iconv}
    (x^i_{carrier}, x^i_{secret}) = iConv(a^i_{carrier}, a^i_{secret})
    ,
\end{equation}
where $iConv(\cdot)$ indicates the forward procedure of invertible convolution.
After that, the output of this ACB can be formulated as:
\begin{equation}\label{eq:acb}
    \begin{split}
        a^{i + 1}_{carrier} &= x^i_{carrier} + \phi(x^i_{secret}), \\
        a^{i + 1}_{secret} &= x^i_{secret} \odot exp(\rho(a^{i + 1}_{carrier})) + \
        \eta(a^{i + 1}_{carrier})
        ,
    \end{split}
\end{equation}
in which $exp(\cdot)$ is the exponential function, $\odot$ is the Hadamard product and 
$\phi(\cdot)$, $\rho(\cdot)$, $\eta(\cdot)$ are functions learned by convolution blocks. 
There can be a variety of choices for the type of these convolution blocks, in this paper,
we use dense block \cite{wang2018esrgan} due to its powerful representation ability. 
In the revealing procedure, the output of the 
$i$\textsuperscript{th} ACB can be obtained by reformulating \autoref{eq:iconv} and \autoref{eq:acb} like:
\begin{equation}
    (a^i_{carrier}, a^i_{secret}) = iConv^{-1}(x^i_{carrier}, x^i_{secret})
    ,
\end{equation}\vspace{-5pt}
and
\begin{equation}
    \begin{split}
        x^i_{secret} &= (a^{i + 1}_{secret} - \eta(a^{i + 1}_{carrier})) \odot 
        exp(-\rho(a^{i + 1}_{carrier})), \\
        x^i_{carrier} &= a^{i + 1}_{carrier} - \phi(x^i_{secret})
        ,
    \end{split}
\end{equation}

where $iConv^{-1}(\cdot)$ is the inverse operation of $iConv(\cdot)$.

Since the input and output of ISN are the same, we only extract the first 12 channels 
from the output and perform IWT to obtain the final encoded image. Similarly, in the
revealing process, we perform DWT on the encoded image and pad it with a constant matrix
$I_z$ to match its size to that of the input of ISN.

\begin{figure}[htb]
\centering
\begin{minipage}[t]{0.33\linewidth}
\centering
\includegraphics[width=\linewidth]{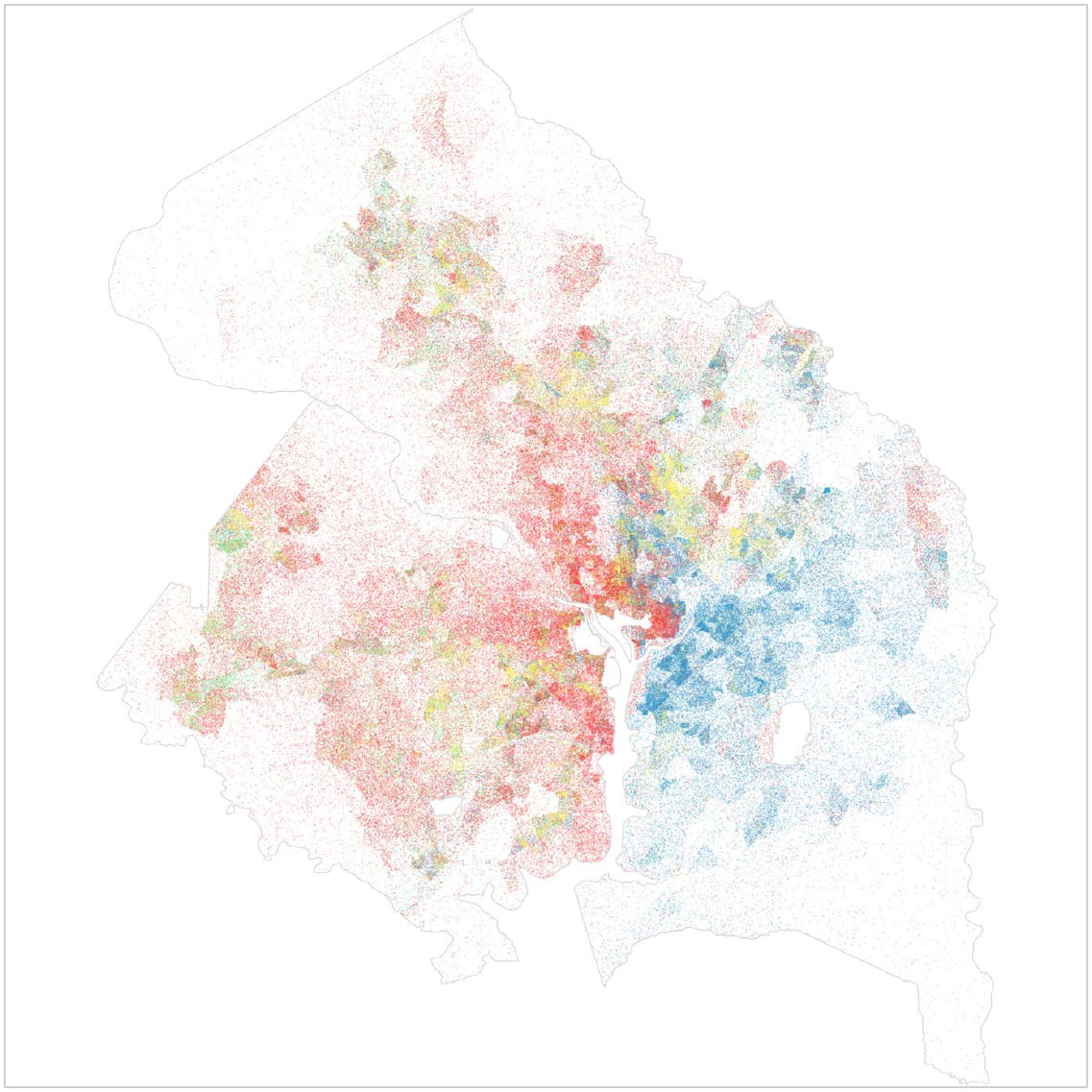}
\subcaption{Original chart}
\end{minipage}\hfill
\begin{minipage}[t]{0.33\linewidth}
\centering
\includegraphics[width=\linewidth]{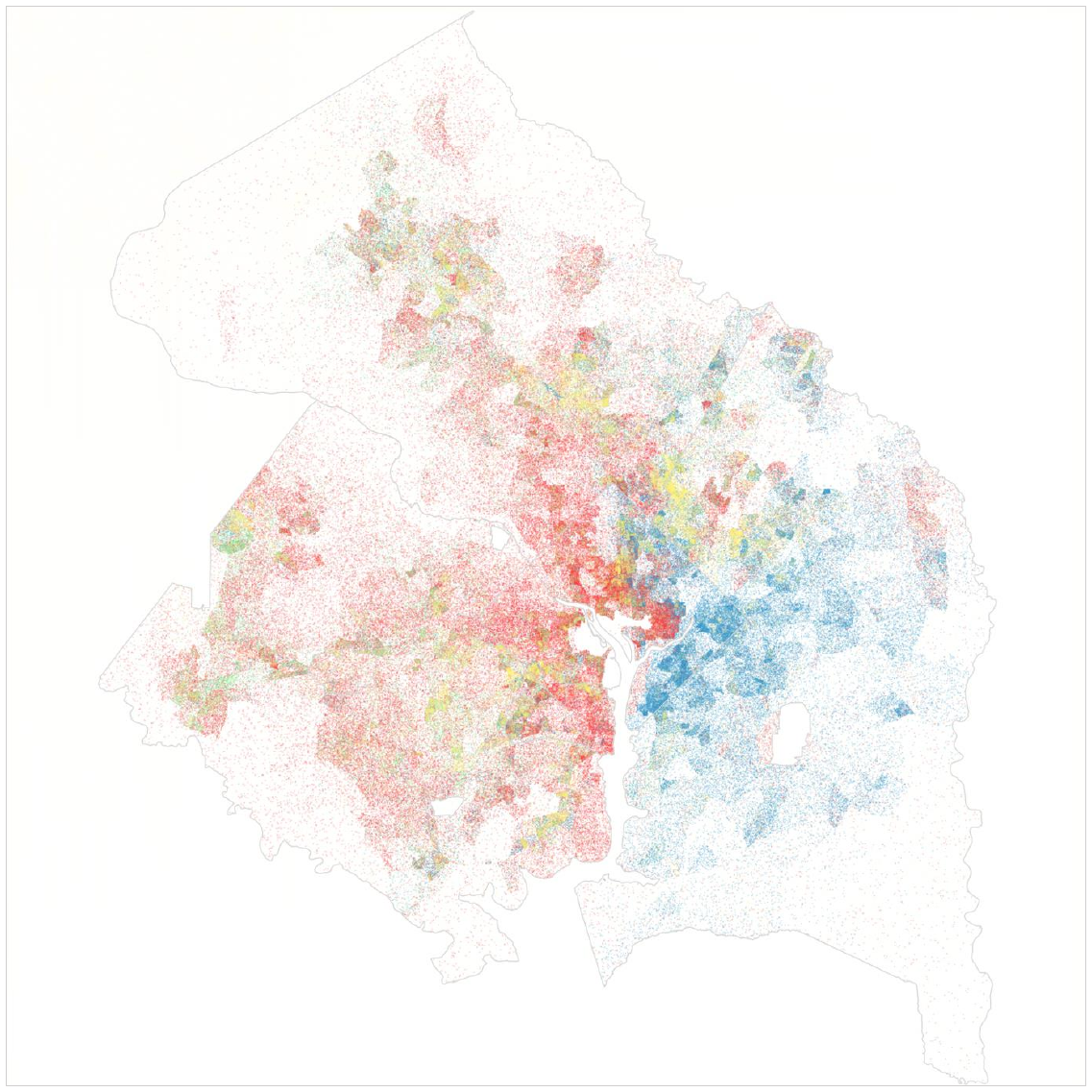}
\subcaption{Encoded Image}
\end{minipage}\hfill
\begin{minipage}[t]{0.33\linewidth}
\centering
\includegraphics[width=\linewidth]{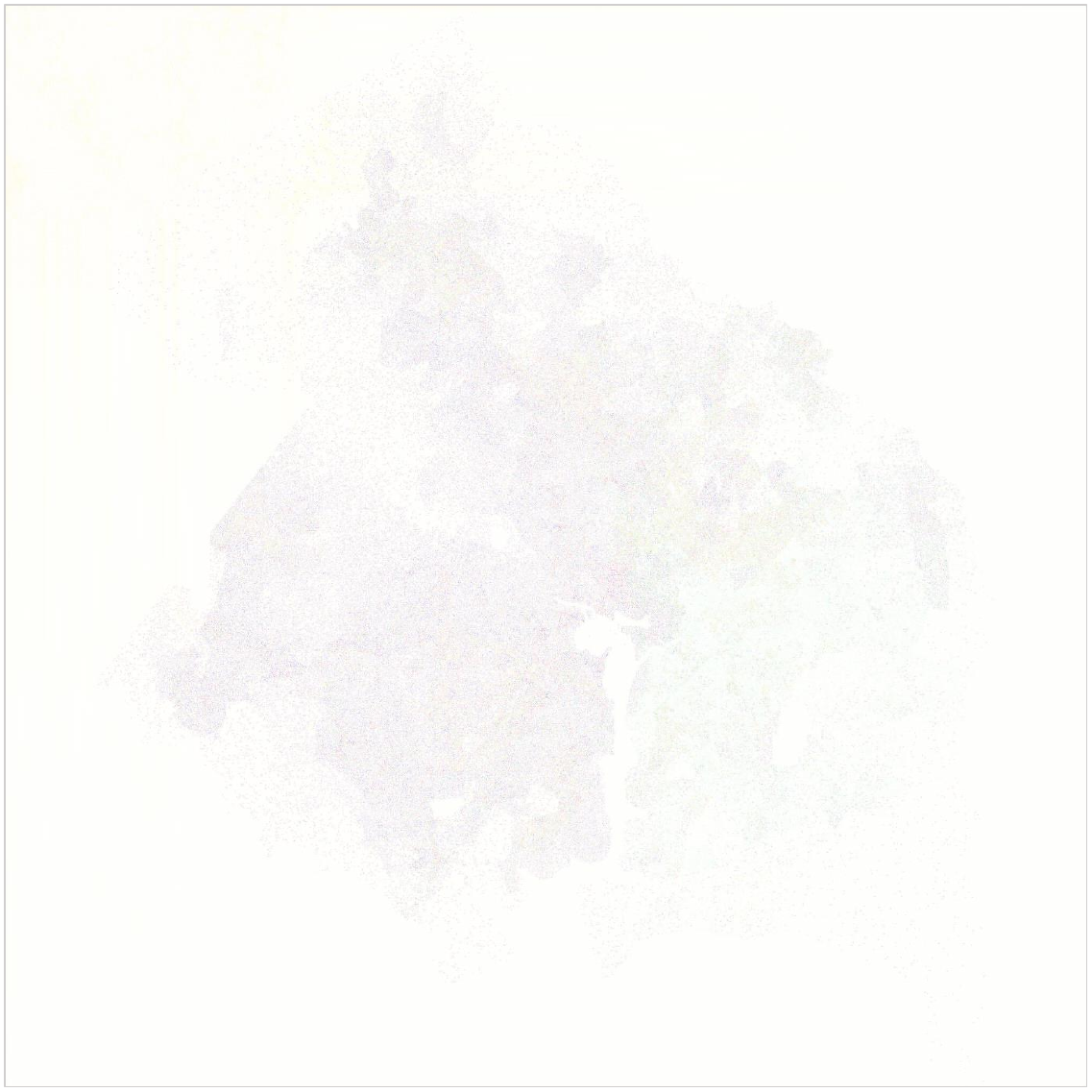}
\subcaption{Reisual $\times$ $2$}
\end{minipage}\vspace{-9pt}
\caption{\label{fig:encode_residual}Difference between the initial chart image and encoded image.}\vspace{-15pt}
\end{figure}

\autoref{fig:encode_residual} demonstrates a comparison of the image quality before and 
after data embedding. \autoref{fig:encode_residual}(c) shows the difference between
the original chart image and the encoded image. We enhance the residual by 2 times to 
more clearly show the difference. As we can see, our method can embed data 
into images while preserving the perceptual quality of the image.

\subsubsection{Loss Function}
The optimization of our network can be divided into two parts, which are the perceptual
quality of the encoded image and the restoration accuracy of secret images. We utilize hybrid
loss functions to train our model in order to achieve better performance.

Given a carrier image $I_{carrier}$ and an encoded image $I_{enc}$, we define our 
encoding loss $\mathcal{L}_{enc}$ as:
\begin{equation}
    \mathcal{L}_{enc} = \mathcal{L}_{mse} + \alpha\mathcal{L}_{freq}
    ,
\end{equation}
where $\alpha$ is a weight coefficient and $\mathcal{L}_{mse}$ measures the mean squared error (MSE) of pixel values between 
$I_{carrier}$ and $I_{enc}$. Inspired by Jing et al. \cite{jing2021hinet}, we introduce 
$\mathcal{L}_{freq}$ to measure and optimize the low-frequency sub-bands difference after wavelet 
decomposition. Assume that $W(\cdot)_l$ output the low-frequency sub-bands of image, 
$\mathcal{L}_{freq}$ can be defined as:
\begin{equation}
    \mathcal{L}_{freq} = \mathcal{L}_{mse}(W(I_{carrier})_l, W(I_{enc})_l)
    .
\end{equation}

For restoration accuracy, we use both L1 and MSE loss function to guide the training 
process, the restoration loss $\mathcal{L}_{res}$ of $N$ secret images can be defined as:
\begin{equation}
    \mathcal{L}_{res} = \sum_{i=1}^{N}(\mathcal{L}_{L1}(I_{secret}^{'i}, I_{secret}^i) + 
    \mathcal{L}_{mse}(I_{secret}^{'i}, I_{secret}^i))
    ,
\end{equation}
where $I_{secret}^i$ and $I_{secret}^{'i}$ denotes the $i$\textsuperscript{th} secret image and 
its corresponding restored image respectively.
In summary, the total loss function of our network can be formulated as:
\begin{equation}\label{eq:totloss}
    \mathcal{L}_{total} = \mathcal{L}_{enc} + \beta\mathcal{L}_{res} = 
    \mathcal{L}_{mse} + \alpha\mathcal{L}_{freq} + \beta\mathcal{L}_{res}
    ,
\end{equation}
where $\beta$ is another weight coefficient just like $\alpha$.

\section{Applications}
Our InvVis is competent in various scenarios, such as embedding URL or copyright information in a visualization, 
which has been discussed in \cite{zhang2020viscode}. Compared to previous methods, our approach has the 
advantage of being applicable to scenarios that involve large amounts of data. For instance,
it can facilitate invertible visualization of data-intensive charts, large-scale source code embedding and 
scientific data embedding.

\begin{figure}[htb]
    \centering
    {\includegraphics[width=1.0\linewidth]{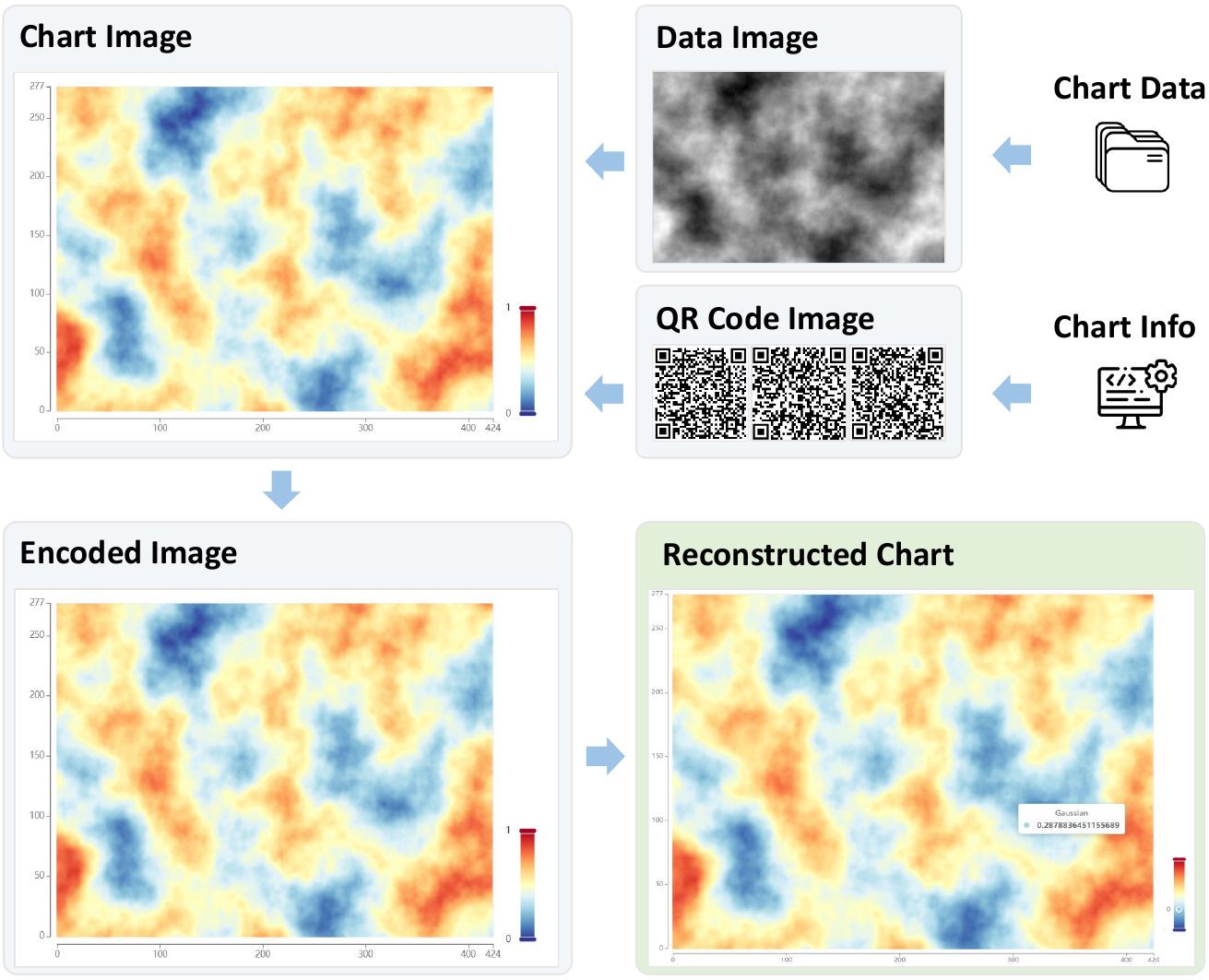}}
    \caption{\label{fig:ui}The application interface of our InvVis. Users can upload the visualization image and data to
    obtain an encoded image. By decoding the image, the visualization can be reconstructed.
    }\vspace{-13pt}
\end{figure}

\subsection{Invertible Visualization of Data-Intensive Chart}
Data-intensive charts are common in the field of information visualization. Due to the large data volume they
contain, such charts generally require some interactive functionalities to assist users in comprehending the 
data. However, there is a lack of a convenient way to disseminate this kind of visualization while preserving their 
interactivity. InvVis enables the reconstruction of interactive charts from a single image, greatly simplifying
their dissemination. \autoref{fig:ui} shows the application interface of our InvVis. In the data embedding process, 
the user can upload a chart image and its metadata, including the chart data and chart information, the application 
can embed the uploaded data into the chart image and output an encoded image. In the data restoring procedure, 
the application can decode an encoded image and reconstruct the interactive chart.

\begin{figure}[htb]
    \centering
    \begin{minipage}[t]{0.48\linewidth}
    \centering
    \includegraphics[width=\linewidth]{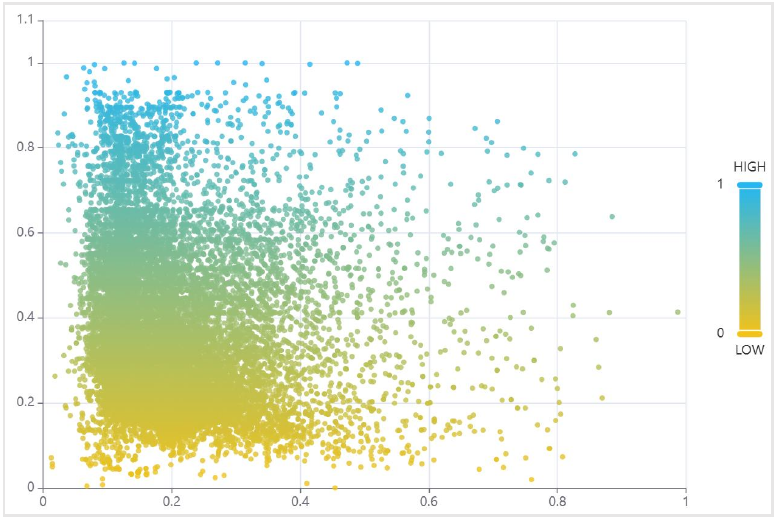}
    \subcaption{Original chart}
    \end{minipage}
    \begin{minipage}[t]{0.48\linewidth}
    \centering
    \includegraphics[width=\linewidth]{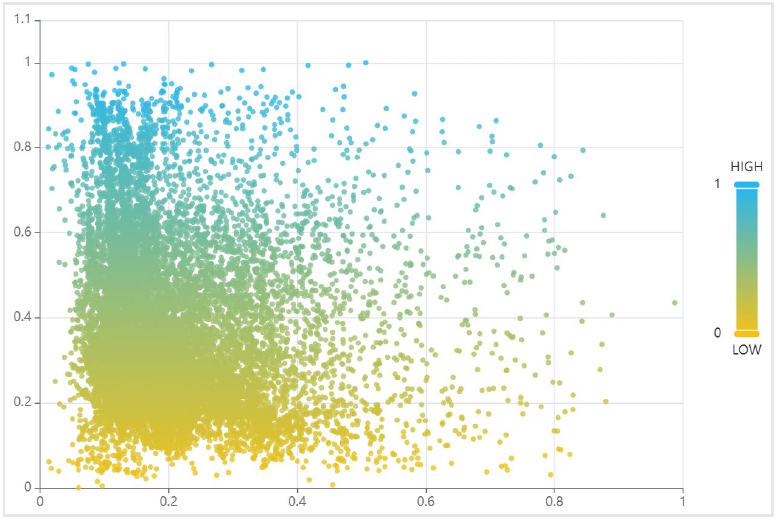}
    \subcaption{Reconstructed chart}
    \end{minipage}
    \hfill
    \begin{minipage}[t]{0.48\linewidth}
    \centering
    \includegraphics[width=\linewidth]{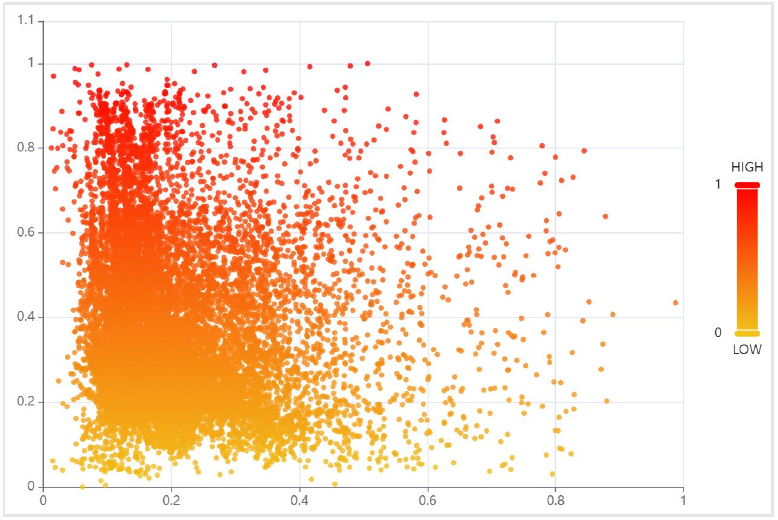}
    \subcaption{Redesigned chart}
    \end{minipage}
    \begin{minipage}[t]{0.48\linewidth}
    \centering
    \includegraphics[width=\linewidth]{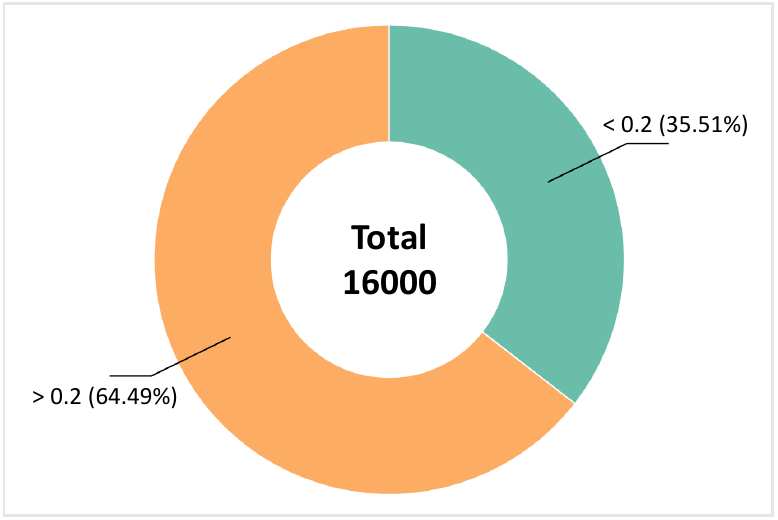}
    \subcaption{Chart using reorganized data}
    \end{minipage}
    \caption{\label{fig:reuse}
    Apart from reconstructing the chart from a single image (b), InvVis can further facilitate the reusing of chart data 
    like redesigning the chart (c) and helping generate a new statistical chart (d).}\vspace{-12pt}
    \end{figure}

InvVis can also facilitate the reusing of chart data. Since it allows to hide all the original data into the 
chart image, users can have full access to all the metadata of the chart after decoding and restoring the data. 
The recovered data contains not only the chart information but also the underlying data, which means that users may 
further reuse the data. As shown in \autoref{fig:reuse}, apart from reconstructing the chart, users may redesign the
chart from various perspectives with the decoded metadata, such as changing the visual 
mapping of the chart. Also, the chart data can be reorganized and mapped to new visualizations. For example, users 
can compute the statistics of the chart data and map it to a doughnut chart.

\subsection{Large-Scale Source Code Embedding}
Source code is an important part of visualizations. Some visualization charts may contain a large amount of source 
code due to their sophisticated interactive methods or large data volume. VisCode \cite{zhang2020viscode} proposed to hide the 
source code in a chart image, but due to the limited steganography capacity of this method, it cannot handle the cases in 
which the amount of source code or data is quite large. As will be discussed in \ref{eva: cap}, InvVis has a much larger 
steganography capacity than VisCode, even when only hiding QR Codes, 
which can be indicated by \autoref{tab:stegCap1}.
Therefore, our method can better handle this problem.

To better illustrate the practicality of large-scale source code embedding, we present an application scenario, 
which is the visualization dashboard. Visualization dashboard is a collection of various visualization charts 
that are displayed together on a single screen or interface. It can be used in business intelligence or data 
analysis to provide a quick overview of key metrics and statistical data. Since a visualization dashboard generally
contains several visualization charts, the amount of its metadata may be much larger than that of a single chart.
Additionally, if the chart data is also written directly in the source code, it will make the source code very long.
For methods like VisCode that encode all the source code into QR Codes, the number of QR Codes can be too large to be hidden in the 
carrier image. Meanwhile, if a relatively small image is used to display a visualization dashboard, the charts and data 
contained in it may be too small to be clearly seen. Therefore, if the source code, together with the chart data of a visualization 
dashboard can be effectively hidden in an image, it can greatly facilitate the dissemination of this kind of visualization.

\begin{figure}[htb]
    \centering
    {\includegraphics[width=0.96\linewidth]{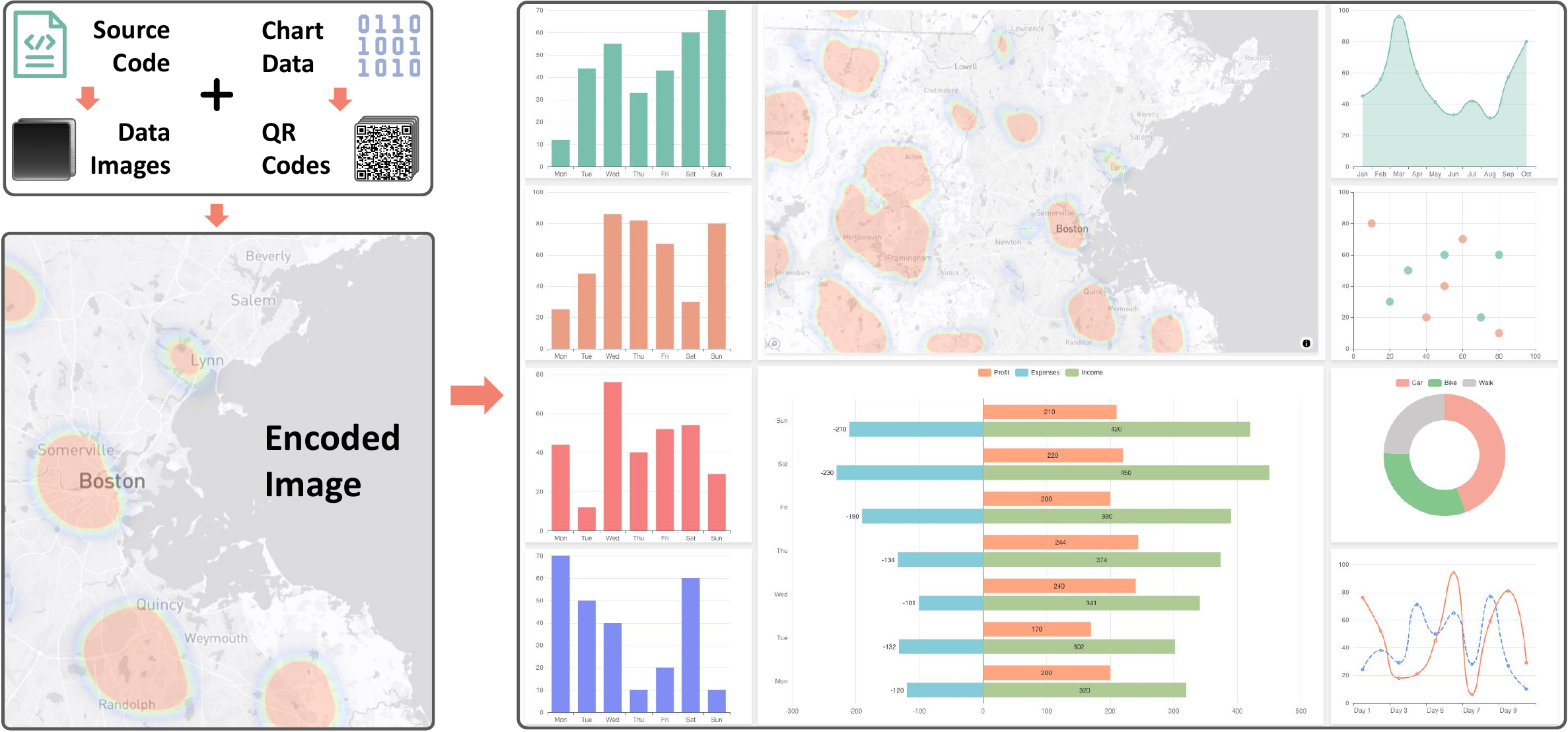}}
    \caption{\label{fig: dashboard}
        An interactive visualization dashboard can be reconstructed by decoding the encoded image.
    }\vspace{-15pt}
\end{figure}

InvVis can handle this problem in a novel way. Since InvVis can hide both data images and QR Code image simultaneously, we can separate 
the data with large volume from the source code and transfer them to data images. This effectively reduces the length of the source code, 
and meanwhile increases the efficiency of data embedding. Our method allows both the data and source code to be carried by a relatively 
smaller carrier image. \autoref{fig: dashboard} demonstrates an example, the source data of the heat map is separated from the 
source code and transferred into data images. The data images and the QR Code image that contains the source code are embedded into 
a relatively small image. By decoding this image, users can obtain a complete and interactive visualization dashboard.

\subsection{Scientific Data Embedding}
Scientific data typically has large data volume, such as terrain data, ocean current data, volumetric data, etc.
These kinds of data generally require visualization methods to present the data and design some interactive ways to 
facilitate researchers or common users in understanding the data intuitively. Since our InvVis can effectively embed 
data into carrier images through data images, it has great potential for the embedding of scientific data. This 
allows users to spread scientific visualizations through a single image.

\begin{figure}[htb]
    \centering
    \begin{minipage}[t]{0.49\linewidth}
    \centering
    \includegraphics[width=\linewidth]{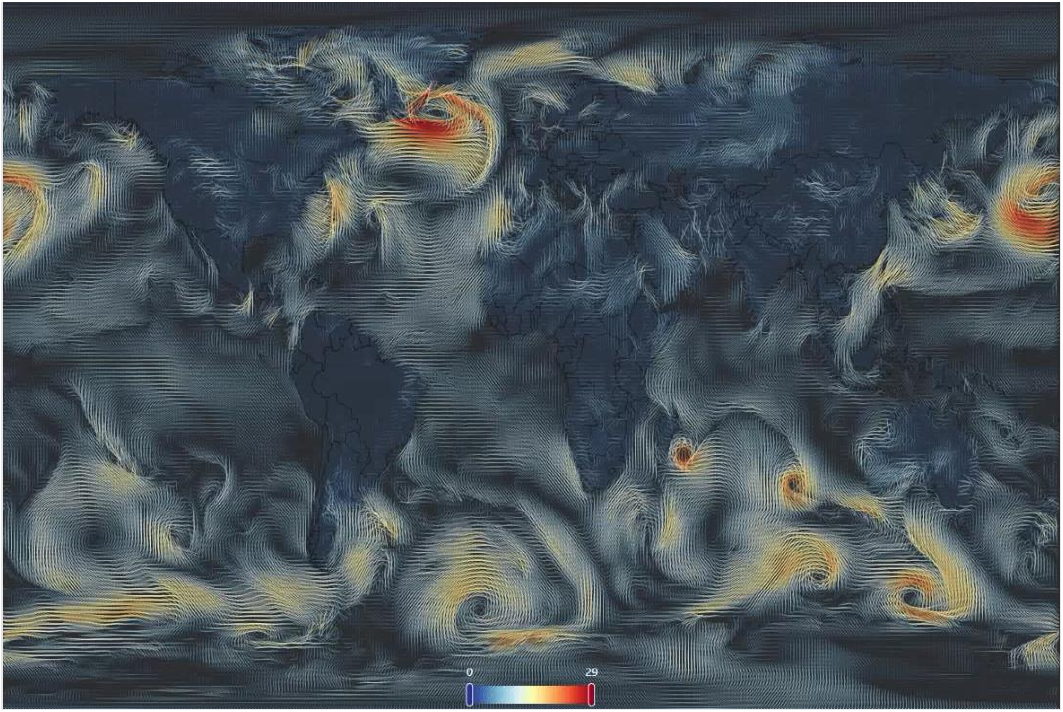}
    \vspace{-10pt}
    \renewcommand{\thesubfigure}{a}
    \subcaption{Image encoded with wind field data}
    \end{minipage}
    \hfill
    \begin{minipage}[t]{0.49\linewidth}
    \centering
    \includegraphics[width=\linewidth]{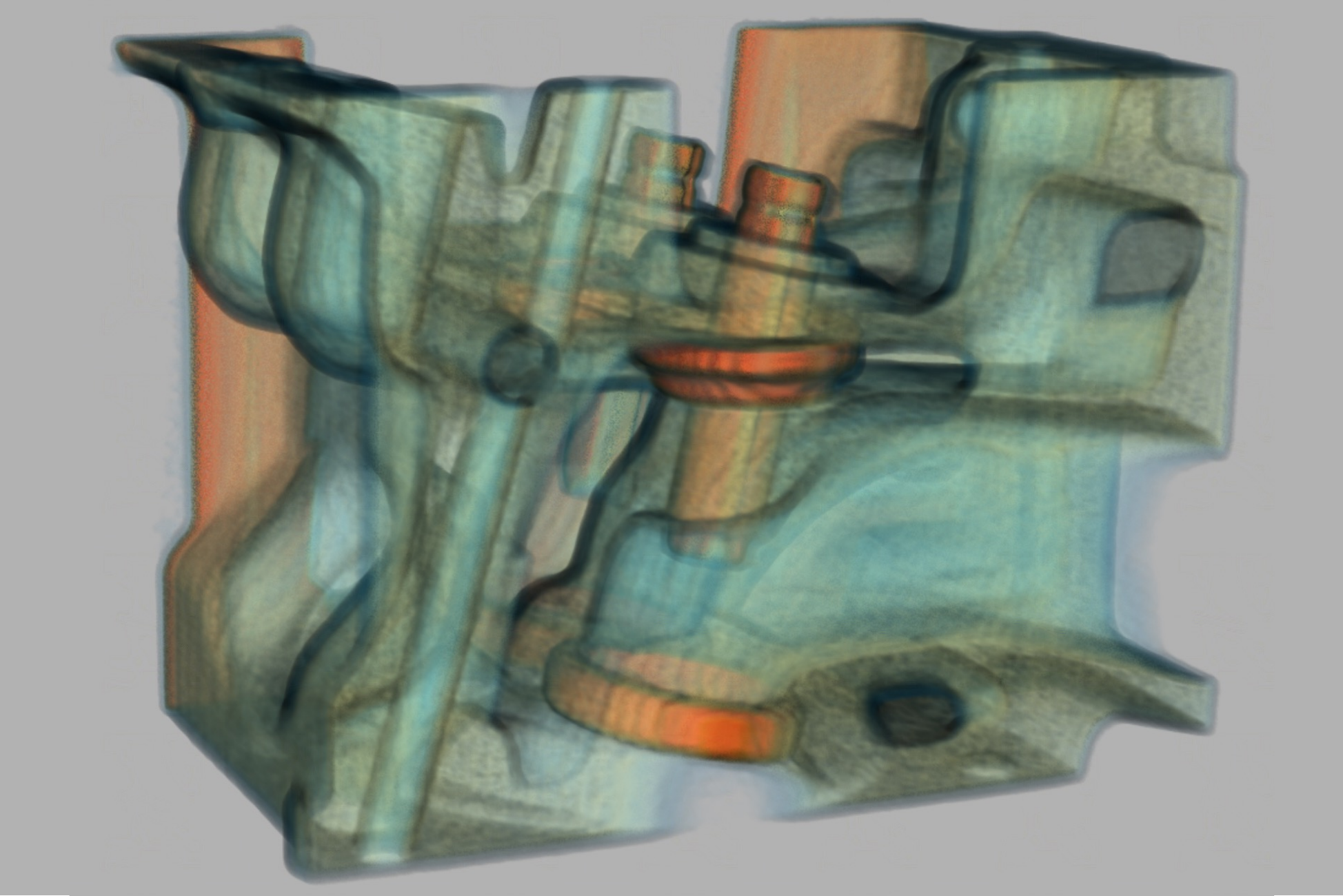}
    \vspace{-10pt}
    \renewcommand{\thesubfigure}{d}
    \subcaption{Image encoded with volumetric data}
    \end{minipage}
    \\
    \begin{minipage}[t]{0.49\linewidth}
    \centering
    \includegraphics[width=\linewidth]{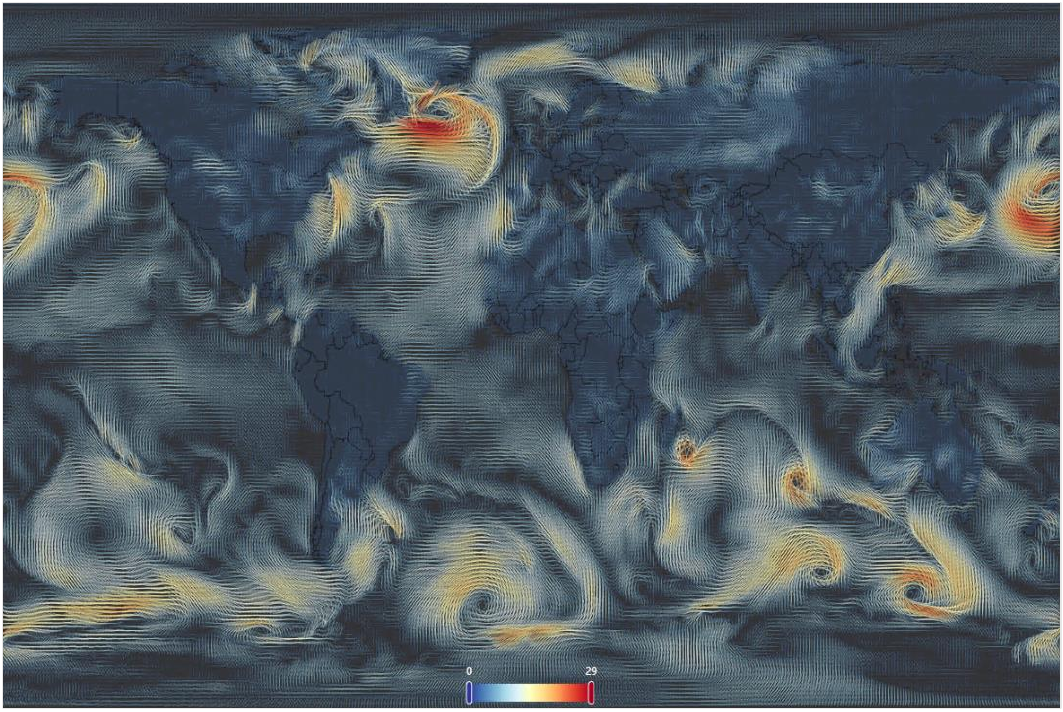}
    \vspace{-10pt}
    \renewcommand{\thesubfigure}{b}
    \subcaption{Reconstructed visualization}
    \end{minipage}
    \hfill
    \begin{minipage}[t]{0.49\linewidth}
    \centering
    \includegraphics[width=\linewidth]{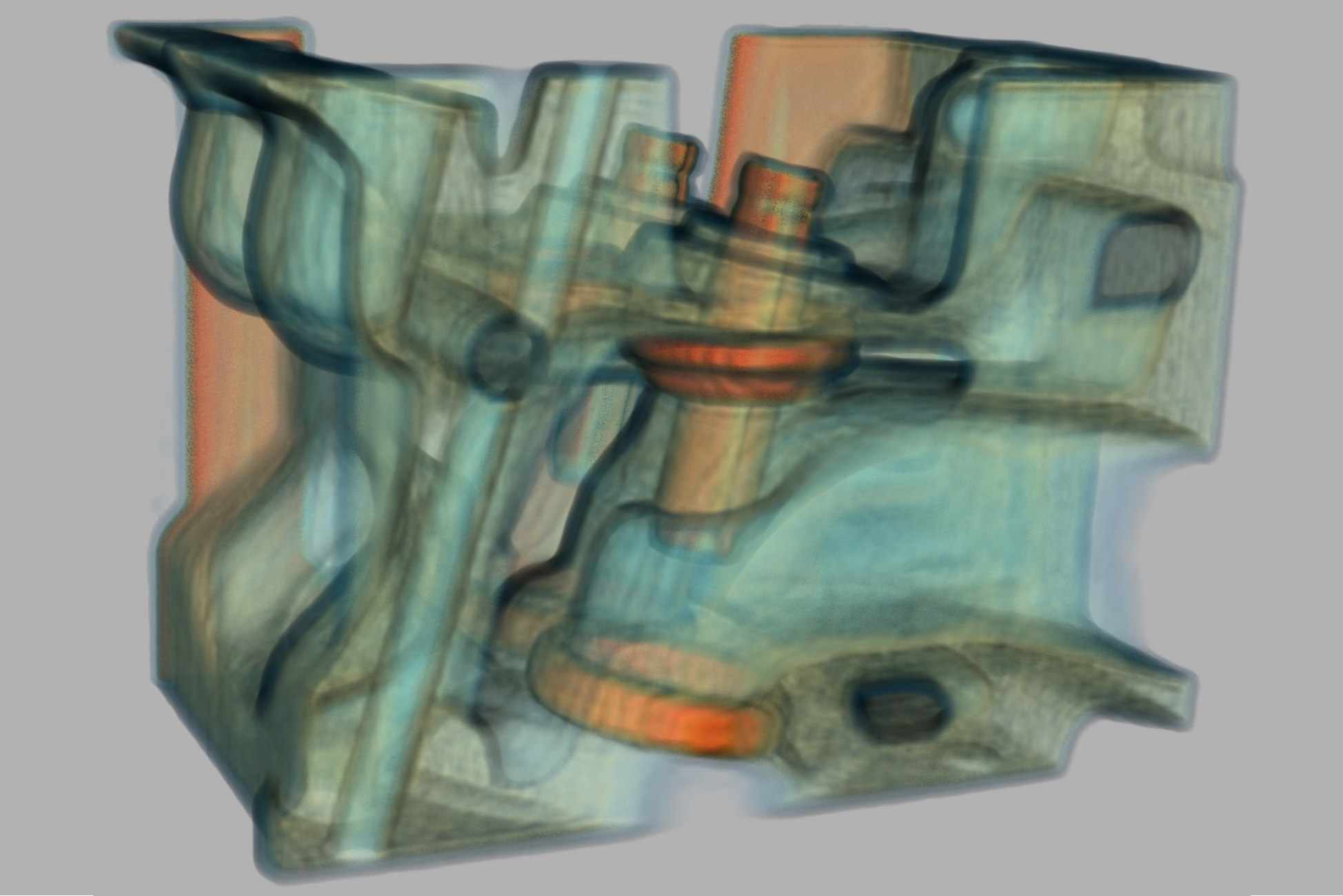}
    \vspace{-10pt}
    \renewcommand{\thesubfigure}{e}
    \subcaption{Reconstructed visualization}
    \end{minipage}
    \\
    \begin{minipage}[t]{0.49\linewidth}
        \centering
        \includegraphics[width=\linewidth]{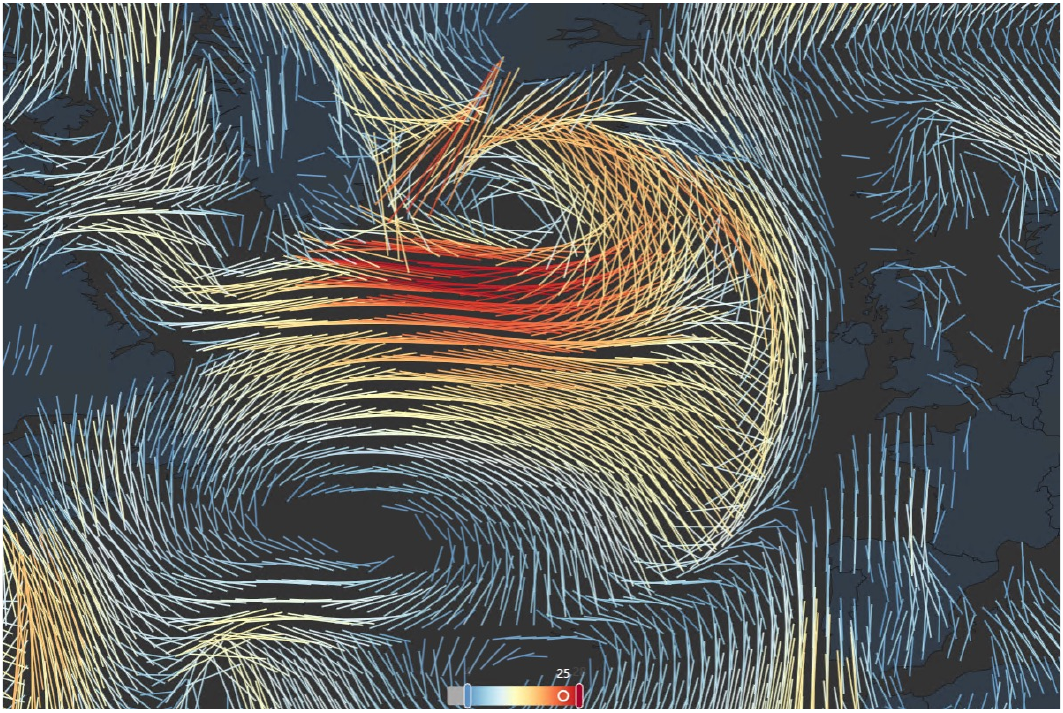}
        \vspace{-10pt}
        \renewcommand{\thesubfigure}{c}
        \subcaption{The reconstructed visualization scaled by user}
    \end{minipage}
    \hfill
    \begin{minipage}[t]{0.49\linewidth}
        \centering
        \includegraphics[width=\linewidth]{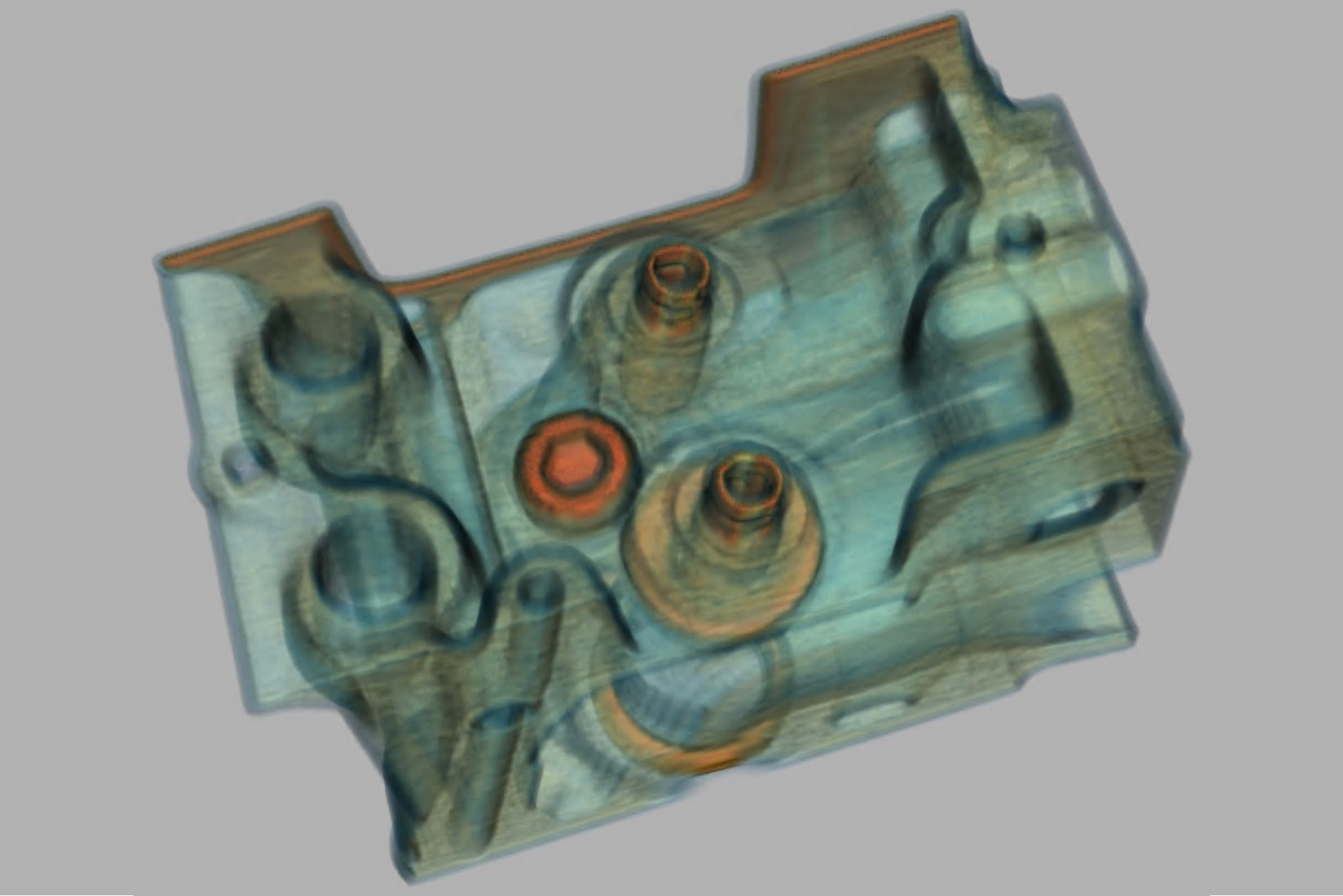}
        \vspace{-10pt}
        \renewcommand{\thesubfigure}{f}
        \subcaption{The reconstructed visualization rotated by user}
    \end{minipage}\vspace{-8pt}
    
    \caption{\label{fig:sci}
    The encoded visualization images and the reconstructed visualizations after decoding.
    }\vspace{-15pt}
\end{figure}

\autoref{fig:sci}(a)(b)(c) shows an example of scientific data embedding. We embed a set of wind field
data which contains a total of 35180 2-dimensional vectors into an image. Although the data volume is quite large, the
size of the carrier image is as small as 520 $\times$ 360. \autoref{fig:sci}(b)(c) shows the reconstructed visualization, which 
is interactive, allowing users to query data within a certain range or zoom in to view the distribution of data in 
a specific area. \autoref{fig:sci}(d)(e)(f) demonstrates another example, we embed the volumetric data of an engine
into a carrier image. The amount of this volumetric data is very large, reaching 256 $\times$ 256 $\times$ 128, while 
in contrast, the carrier image is only 2304 $\times$ 1280. Our method can effectively hide such a large amount of data 
into the carrier image and results \autoref{fig:sci}(d), whose distortion is almost unobservable. And the restored 
data is also very accurate, as shown in \autoref{fig:sci}(e)(f), the reconstructed visualization is indistinguishable 
from the original one and allows users to view the volume data visualization from different directions.

\section{Evaluation}
We evaluate our method from two aspects, steganography quality and steganography capacity,
which are two of the most important criteria for information steganography.
We conduct our experiments on a PC with an Intel Core i7 CPU, an NVIDIA GeForce 3090 
GPU, and 64 GB of memory. Our model is implemented with PyTorch \cite{paszke2019pytorch}.
We use 32 ACBs in our ISN and train our model using the Adam optimizer \cite{Kingma2014AdamAM}. 
The weight coefficients of the loss function (\autoref{eq:totloss}) are set as $\alpha = 0.5$ and $\beta = 1.6$
to balance the data embedding quality and data restoration accuracy.
In addition, as mentioned in \autoref{sec:dataset}, our experiments are conducted on our
test dataset.

\subsection{Steganography Quality}
\label{sec:steg_q}
Steganography quality refers to both the perceptual quality of the encoded image and the 
accuracy of restored information. These two criteria are inversely related to each
other, which means higher embedding quality generally leads to lower data restoration
capacity and vice versa. A good steganography method should achieve a satisfactory 
trade-off between them. 

We evaluate the quality of encoded image using three metrics, which are the peak 
signal-to-noise ratio (PSNR) \cite{almohammad2010stego}, the structural similarity 
index (SSIM) \cite{wang2004image} and the learned perceptual image patch similarity 
(LPIPS) \cite{zhang2018unreasonable}. The PSNR is often used to evaluate the
distortion of images, and the SSIM measures the structural similarity between two 
images. The LPIPS is based on the high-level features of images, which represent 
the perceptual similarity between two images. In our experiments, we use VGG 
\cite{DBLP:journals/corr/SimonyanZ14a} as the feature extraction network for the calculation of LPIPS.
Higher PSNR, SSIM and lower LPIPS is better.

For the data restoration accuracy, we use root mean squared error (RMSE) to measure 
the difference between the initial data images and the restored data images. In addition, we 
use the text recovery accuracy (TRA) \cite{zhang2020viscode}, which refers to the 
proportion of the restored characters among all input characters, to measure the 
performance of QR Code image recovery. Lower RMSE and higher TRA is better.

\begin{figure}[htb]
    \centering
    \begin{minipage}[t]{0.48\linewidth}
    \centering
    \includegraphics[width=\linewidth]{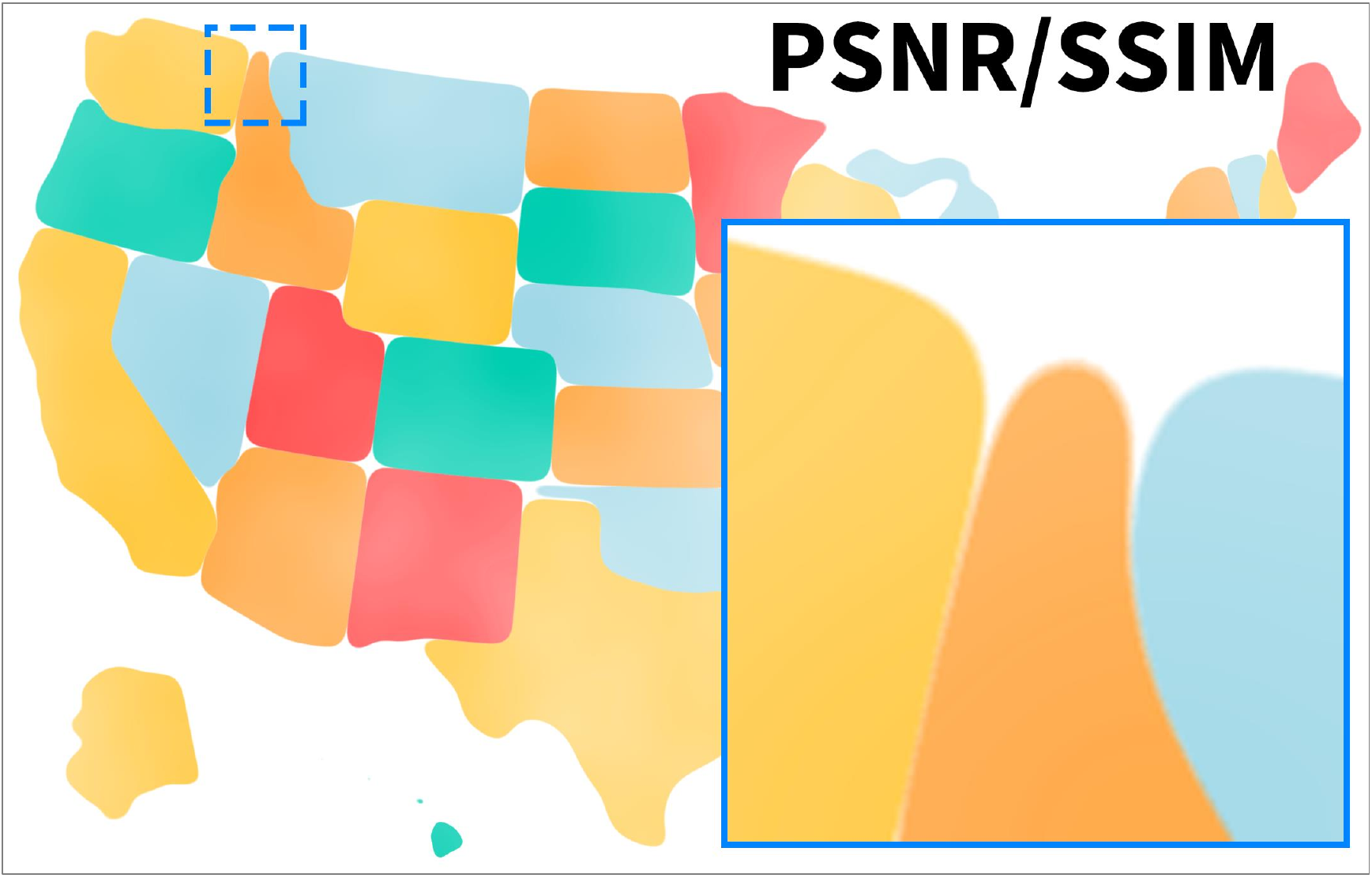}
    \subcaption{Original chart}
    \end{minipage}
    \begin{minipage}[t]{0.48\linewidth}
    \centering
    \includegraphics[width=\linewidth]{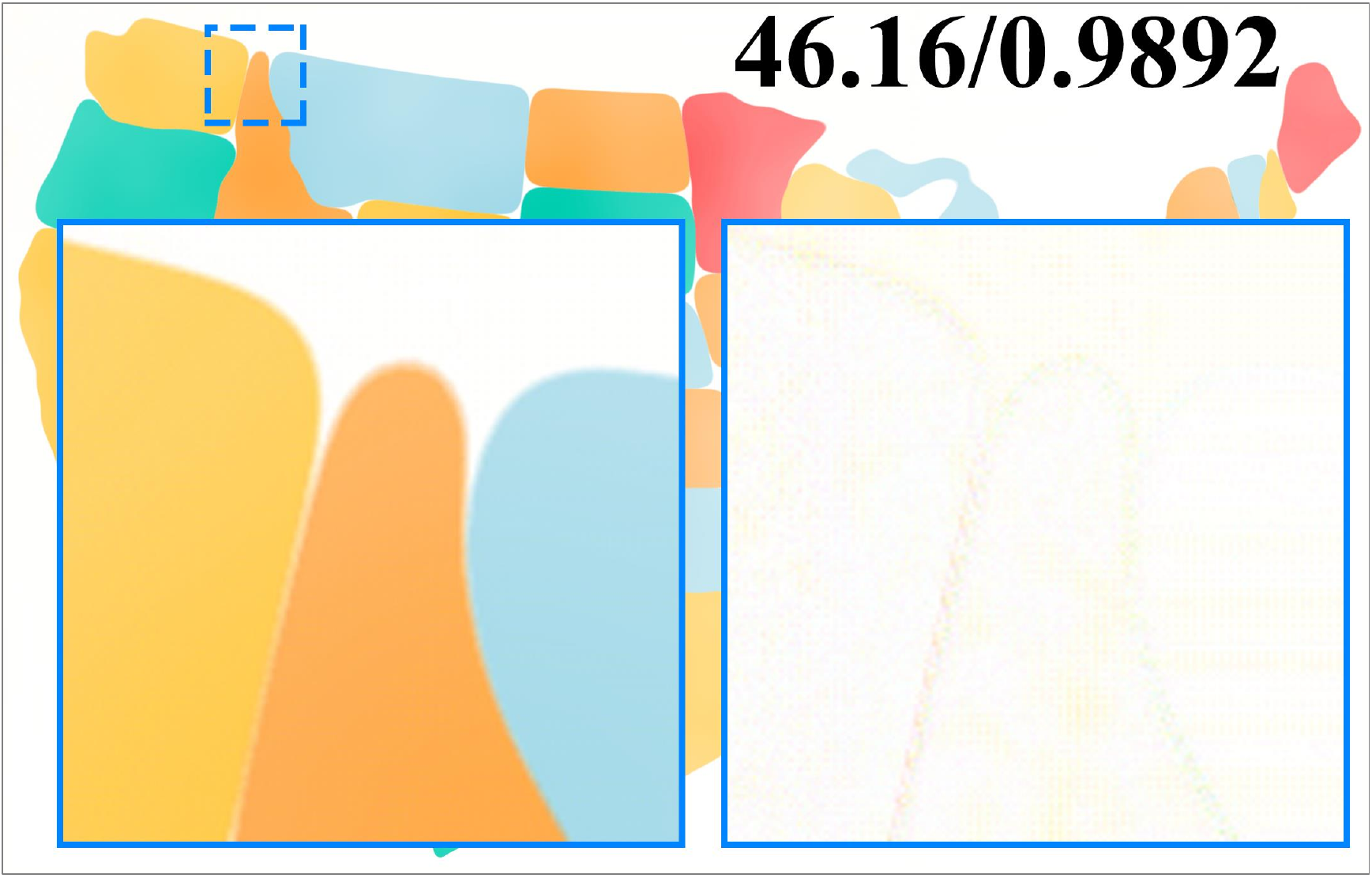}
    \subcaption{Ours}
    \end{minipage}
    \hfill
    \begin{minipage}[t]{0.48\linewidth}
    \centering
    \includegraphics[width=\linewidth]{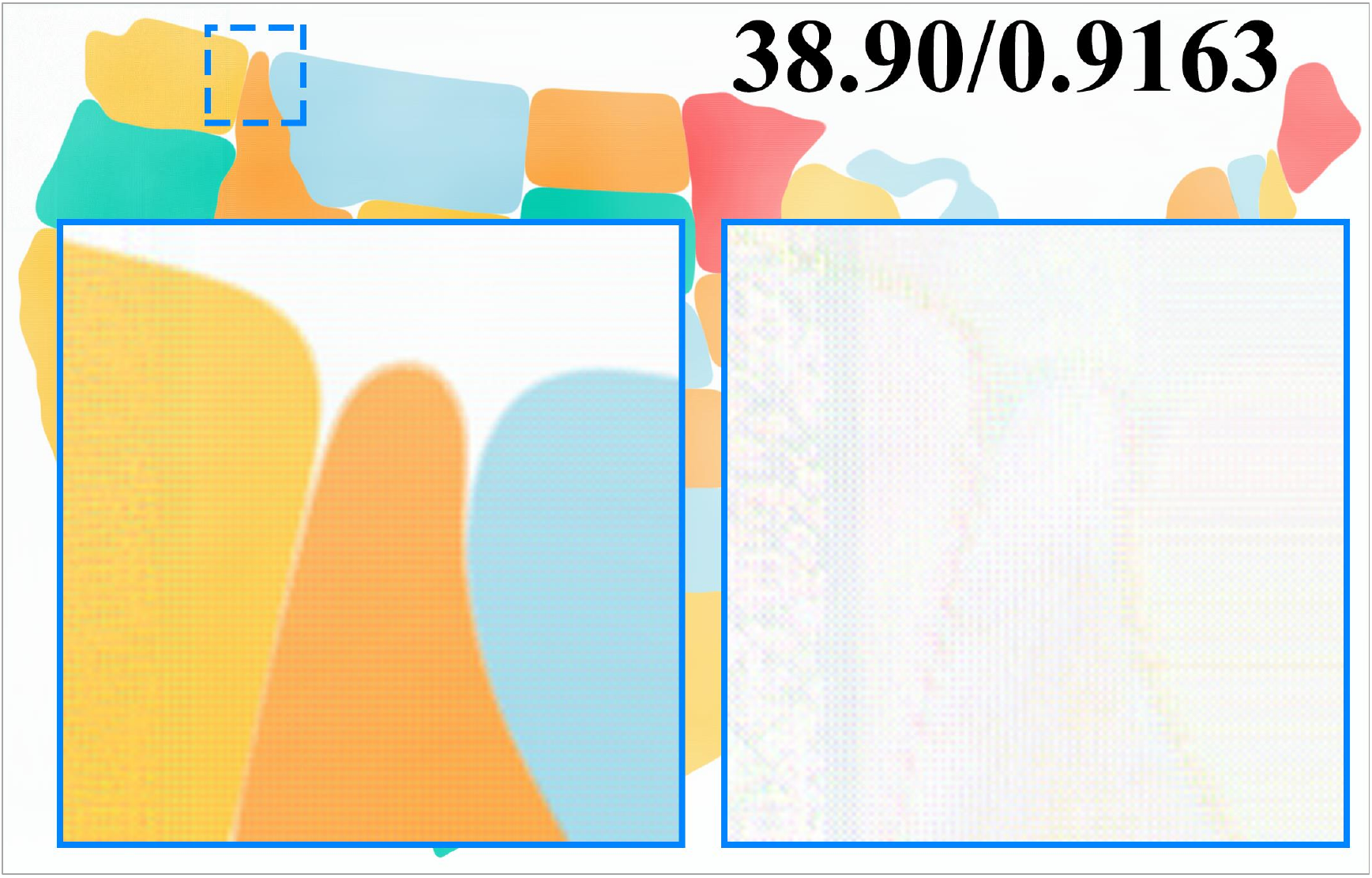}
    \subcaption{HiNet}
    \end{minipage}
    \begin{minipage}[t]{0.48\linewidth}
    \centering
    \includegraphics[width=\linewidth]{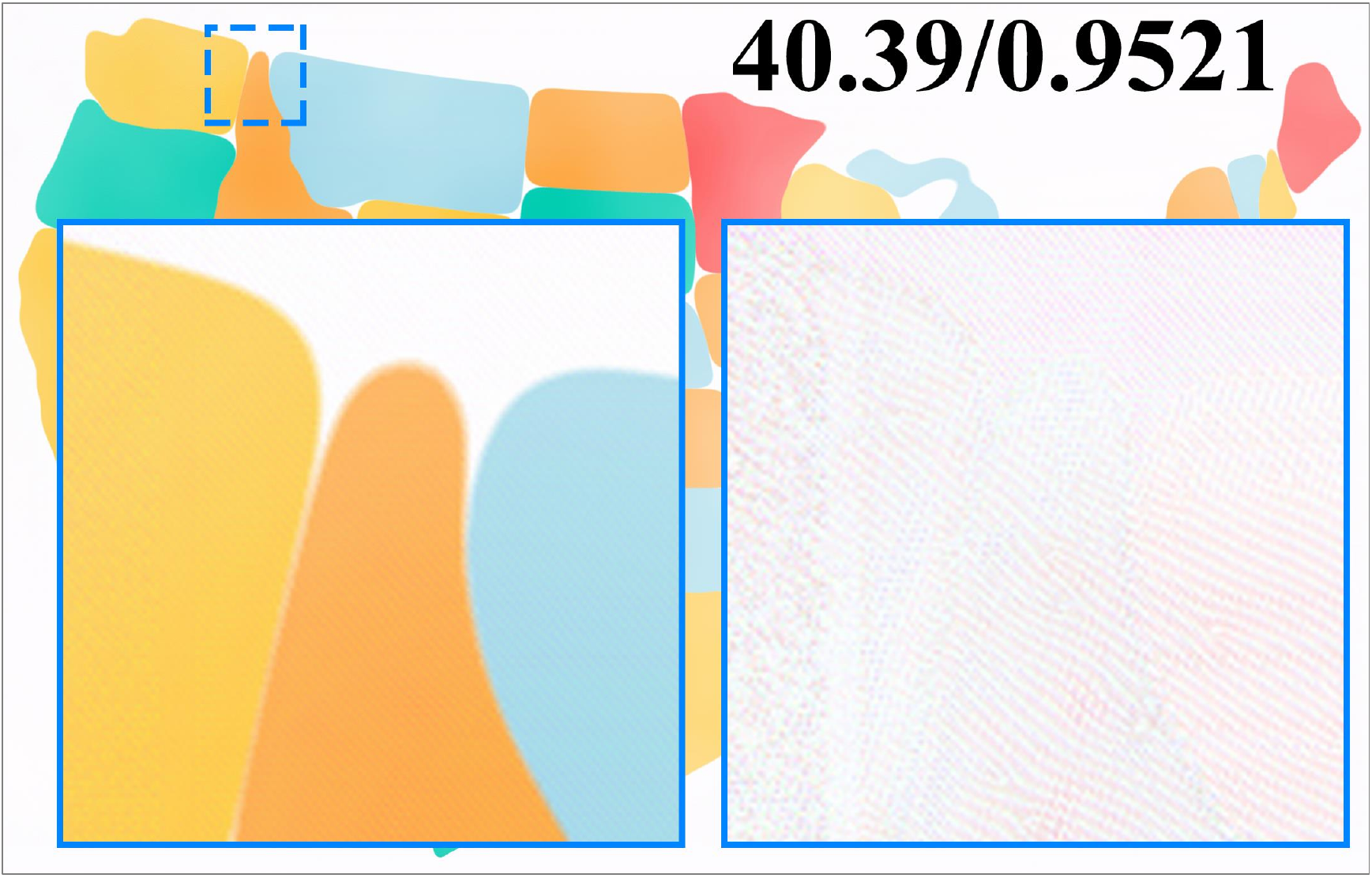}
    \subcaption{IICNet}
    \end{minipage}\vspace{-10pt}
    \caption{\label{fig:encode_compare}
    Comparison of the encoded images with other methods. The residual is enhanced by 4 times to clearly show the difference.}\vspace{-10pt}
    \vspace{-5pt}\end{figure}


We first conduct a series of ablation experiments to investigate the impact of FFB and 
wavelet domain steganography on the final perceptual quality of the encoded image. We hide 3 
channels of $512 \times 512$ data images and one channel of $360 \times 360$ QR Code image 
into each carrier image. Each QR Code image is encoded with $T$ characters and 
$T \in [1, 1200]$. As shown in \autoref{tab:ablation}, our full model achieves
the best balance between information embedding quality and data restoration accuracy.\vspace{-7pt}

\begin{table}[htb]
\caption{Ablation experiment results of different metrics}
\vspace{-8pt}
\newcolumntype{M}[1]{>{\centering\arraybackslash}m{#1}}
\renewcommand\arraystretch{1.2}
\centering
\small
\begin{tabular}{@{}M{4.0cm}M|M{0.9cm}M{0.9cm}M{0.9cm}|M{0.9cm}M{0.9cm}|@{}}
\bottomrule
\multicolumn{2}{|c|}{Method} & {PSNR$\uparrow$} & {SSIM$\uparrow$} & {LPIPS$\downarrow$} & {RMSE$\downarrow$} & {TRA$\uparrow$} \\ [0.5pt]
\hline
\multicolumn{2}{|c|}{Ours w/o wavlet} & {$39.1858$} & {$0.9755$} & {$0.0572$} & {$0.0073$} & {$0.4517$} \\
\multicolumn{2}{|c|}{Ours w/o FFB}    & {$41.6315$} & {$0.9681$} & {$0.0566$} & {$0.0066$} & {$0.9916$} \\
\hline
\multicolumn{2}{|c|}{\textbf{Ours}} & \textbf{44.6889} & \textbf{0.9830} & \textbf{0.0120} & \textbf{0.0062} & \textbf{0.9892} \\ [0.5pt]
\toprule
\end{tabular}
\label{tab:ablation}
\end{table}\vspace{-8pt}

In addition, we experiment on how the linear interpolation discussed in \ref{sec:dtoi_discrete} 
affects the steganography quality. We regenerate the scatter data images in our dataset with different 
values of $K$ and retrain the model, and then we evaluate it on the corresponding test set 
(contains scatter data images only). The result 
is shown in \autoref{tab:linear}. As we can see, larger $K$ can lead to better steganography quality 
at the cost of lower steganography capacity.

\begin{table}[htb]\vspace{-7pt}
\caption{Steganography quality under different values of $K$}\vspace{-9pt}
\newcolumntype{M}[1]{>{\centering\arraybackslash}m{#1}}
\renewcommand\arraystretch{1.2}
\centering
\small
\begin{tabular}{@{}M{4.0cm}M|M{1.0cm}M{1.0cm}M{1.0cm}|M{1.0cm}M{1.0cm}|@{}}
\bottomrule
\multicolumn{2}{|c|}{K value} & {PSNR$\uparrow$} & {SSIM$\uparrow$} & {LPIPS$\downarrow$} & {RMSE$\downarrow$} & {TRA$\uparrow$} \\ [0.5pt]
\hline
\multicolumn{2}{|c|}{0} & {$42.3589$} & {$0.9724$} & {$0.0379$} & {$0.0241$} & {$0.9875$} \\
\multicolumn{2}{|c|}{1} & {$43.2026$} & {$0.9776$} & {$0.0258$} & {$0.0202$} & {$0.9918$} \\
\multicolumn{2}{|c|}{3} & {$45.5204$} & {$0.9880$} & {$0.0061$} & {$0.0090$} & {$1.0000$} \\
\multicolumn{2}{|c|}{7} & {$46.3686$} & {$0.9908$} & {$0.0028$} & {$0.0050$} & {$1.0000$} \\ [0.5pt]
\toprule
\end{tabular}
\vspace{-7pt}
\label{tab:linear}
\end{table}

We also compare our model with other image steganography methods, i.e. HiNet 
\cite{jing2021hinet} and IICNet \cite{cheng2021iicnet}. We hide 2 channels of 
$512 \times 512$ data images and one channel of $360 \times 360$ QR Code image 
into each carrier image. \autoref{fig:encode_compare} shows a comparison between 
the encoded image generated by our method with Hinet and IICNet. The magnified area 
and its corresponding residual (enhanced by 4 times), along with the PSNR and SSIM 
of the encoded images are shown. As we can see, InvVis can generate encoded images 
that are more perceptually similar to the carrier image
and has less distortion. The results of the comparative experiments are shown in 
\autoref{tab:comparision}, in which we can see that our model achieves the best performance
on both steganography quality and data restoration accuracy.\vspace{-6pt}

\begin{table}[htb]
 \vspace{-1pt}
\caption{Comparision of our model with other methods}
\vspace{-6pt}
\newcolumntype{M}[1]{>{\centering\arraybackslash}m{#1}}
\renewcommand\arraystretch{1.2}
\centering
\small
\begin{tabular}{@{}M{4.0cm}M|M{1.0cm}M{1.0cm}M{1.0cm}|M{1.0cm}M{1.0cm}|@{}}
\bottomrule
\multicolumn{2}{|c|}{Method} & {PSNR$\uparrow$} & {SSIM$\uparrow$} & {LPIPS$\downarrow$} & {RMSE$\downarrow$} & {TRA$\uparrow$} \\ [0.5pt]
\hline
\multicolumn{2}{|c|}{HiNet \cite{jing2021hinet}} & {$38.9608$} & {$0.9377$} & {$0.0563$} & {$0.0188$} & {$0.9360$} \\
\multicolumn{2}{|c|}{IICNet \cite{cheng2021iicnet}} & {$38.7422$} & {$0.9276$} & {$0.0474$} & {$0.0105$} & {$0.6069$} \\
\hline
\multicolumn{2}{|c|}{\textbf{Ours}} & \textbf{45.5246} & \textbf{0.9866} & \textbf{0.0090} & \textbf{0.0062} & \textbf{0.9961} \\ [0.5pt]
\toprule
\end{tabular}
\vspace{-13pt}
\label{tab:comparision}
\end{table}

\subsection{Steganography Capacity}
\label{eva: cap}
The capacity of steganography concerns how much chart information can be concealed in the image, 
which is crucial for the invertible visualization of charts with large amounts of data. There
is also a trade-off between steganography capacity and quality, larger steganography capacity
can lead to lower quality of the encoded image.

To measure the steganography capacity of different methods, we use bits per pixel (BPP),
which indicates the average number of bits hidden in the encoded image per pixel, as the metric.
Formally, given a carrier image $I$ whose size is $(C, H, W)$, its BPP is calculated as:\vspace{-7pt}
\begin{equation}
    BPP(I) = \frac{L}{C \times H \times W}
    ,\vspace{-5pt}
\end{equation}
where $L$ is the total number of bits hidden in $I$.

We compare our method with VisCode \cite{zhang2020viscode}. VisCode uses a region proposal algorithm 
based on saliency detection to choose the places to hide QR Code images. Although this method can 
achieve better image encoding quality than using predetermined regions for data embedding, it cannot 
guarantee that the image is maximally utilized to hide data. Also, if the encoded data volume is large,
VisCode cannot achieve a satisfactory embedding quality. In contrast, our method concatenates the QR 
codes horizontally and vertically into QR Code images to maximize spatial utilization, and our method can achieve 
good steganography quality even if the data volume is
 large. 

We first compare the steganography capacity of our method with VisCode when only hiding QR Codes. Since 
our methods can not only hide QR Code image but also hide data images, we also compare the maximum 
steganography capacity of our method with VisCode. 
The result is shown in \autoref{tab:stegCap1}, as we can see, our method achieves a larger steganography
capacity than VisCode even when only hiding QR Code image, and our maximum steganography capacity far
exceeds that of VisCode. Moreover, the quality of the encoded image is also much better than that of VisCode
in both experiments. 
Besides, the average computational time of encoding and decoding, together with the
model size of VisCode and InvVis are also shown in \autoref{tab:stegCap1}. As we can see, InvVis has a smaller
model size and faster computational speed.

\begin{table}[htb]
\vspace{-4pt}
\caption{Steganography capacity compared with VisCode}
\vspace{-7pt}
\newcolumntype{M}[1]{>{\centering\arraybackslash}m{#1}}
\renewcommand\arraystretch{1.2}
\centering
\small
\begin{tabular}{|M{1.8cm}|M{0.75cm}|M{0.7cm}|M{0.65cm}M{0.65cm}M{0.65cm}|M{0.8cm}|@{}} 
\bottomrule
{Method} & {Time} & {Size} & {PSNR$\uparrow$} & {SSIM$\uparrow$} & {LPIPS$\downarrow$} & {BPP$\uparrow$} \\ [0.5pt]
\hline
VisCode \cite{zhang2020viscode} & {4.9/2.6} & {$335$mb} & {29.8098} & {0.9499} & {0.0682} & {0.0049} \\
\hline
Ours (QR only) & \multirow{2}{*}{\textbf{3.0/2.2}} & \multirow{2}{*}{\textbf{158mb}} & \textbf{45.1067} & \textbf{0.9851} & \textbf{0.0073} & \textbf{0.0218} \\
Ours (maximum) & & & \textbf{44.5082} & \textbf{0.9822} & \textbf{0.0131} & \textbf{5.5218} \\ [0.5pt]
\toprule
\end{tabular}
\vspace{-5pt}
\label{tab:stegCap1}
\end{table}
    
We also compare the steganography quality of our method under different BPP values.
We divide the BPP into three levels, which are low, medium and maximum. The 
BPP value of different levels and the corresponding steganography quality are shown in 
\autoref{tab:stegCap2}, as we can see, InvVis can maintain high steganography quality under
different levels of BPP.

\begin{table}[htb]
\vspace{-7pt}
\caption{Steganography quality under different BPP levels}
\vspace{-7pt}
\newcolumntype{M}[1]{>{\centering\arraybackslash}m{#1}}
\renewcommand\arraystretch{1.2}
\centering
\small
\begin{tabular}{@{}M{4.0cm}M|M{1.0cm}|M{1.2cm}M{1.2cm}M{1.2cm}|@{}}
\bottomrule
\multicolumn{2}{|c|}{BPP Level} & {BPP} & {PSNR$\uparrow$} & {SSIM$\uparrow$} & {LPIPS$\downarrow$} \\ [0.5pt]
\hline
\multicolumn{2}{|c|}{Low} & {0.5458} & {46.0429} & {0.9893} & {0.0051} \\
\multicolumn{2}{|c|}{Medium} & {2.7604} & {45.5163} & {0.9868} & {0.0104} \\
\multicolumn{2}{|c|}{Maximum} & {5.5218} & {44.5082} & {0.9822} & {0.0131} \\ [0.5pt]
\toprule
\end{tabular}\vspace{-8pt}
\label{tab:stegCap2}
\end{table}\vspace{-5pt}
\section{Conclusion}
We propose InvVis, a new approach for embedding information into visualization images, which is suitable for cases 
with large amounts of data. Our InvVis can support various application scenarios like invertible visualization of
data-intensive charts, large-scale source code embedding, scientific data embedding, etc. We propose a new method 
to transfer chart data into images that can be more easily embedded in chart image with high quality. We also 
outline a deep learning-based pipeline to embed information into chart images. Our evaluation experiments show 
that InvVis can achieve high-quality information concealing and revealing with large steganography capacity.

The current version of InvVis has certain limitations in addressing invertible visualization problems with zero 
tolerance for errors. In addition, the steganography capacity of InvVis is related to the size of the carrier image,
it may not achieve a satisfactory embedding capacity if the image size is too small. Also, InvVis is mainly
designed to facilitate chart image transmission on the internet, so it lacks stability when facing real-world 
image interference such as printing and shooting.

In the future, we plan to propose a more widely applicable method that can handle more types of data and achieve 
better steganography performance and recovery accuracy while ensuring steganography capacity. Additionally, as 
shown in the work of Tancik et al. \cite{tancik2020stegastamp} and Fu et al. \cite{fu2022chartstamp}, the method 
can be more robust if it is stable against some steganography attacks, this is also a potential research direction.

\acknowledgments{
  The authors wish to acknowledge the support from NSFC under Grants (No. 62102152, 62072183 and 61802128), Shanghai Municipal Science and Technology Major Project (Grant No. 22511104600), and CAAI-Huawei MindSpore Open Fund.
}

\bibliographystyle{abbrv-doi-hyperref}

\bibliography{template}


\appendix 

\end{document}